\documentclass[runningheads]{llncs}
\usepackage[T1]{fontenc}
\usepackage{graphicx}
\usepackage{rotating}
\newcommand{\vertcell}[2]{\rotatebox{90}{\begin{minipage}{#1}\begin{center}#2\end{center}\end{minipage}}}
\usepackage{multirow}
\usepackage{longtable}
\usepackage[table]{xcolor}
\begin{document}
% Used for displaying a sample figure. If possible, figure files should
% If you use the hyperref package, please uncomment the following two lines
% to display URLs in blue roman font according to Springer's eBook style:
%\usepackage{color}
%\renewcommand\UrlFont{\color{blue}\rmfamily}
%
%\begin{document}
%
\title{Cross-Domain Robustness of Transformer-based Keyphrase Generation}
\author{Anna Glazkova\inst{1,3}\orcidID{0000-0001-8409-6457} \and
Dmitry Morozov\inst{2,3}\orcidID{0000-0003-4464-1355}}
%\author{First Author\inst{1}\orcidID{0000-1111-2222-3333} \and
%Second Author\inst{2,3}\orcidID{1111-2222-3333-4444} \and
%Third Author\inst{3}\orcidID{2222--3333-4444-5555}}
%
\authorrunning{A. Glazkova, D. Morozov}
%\institute{Princeton University, Princeton NJ 08544, USA \and
%Springer Heidelberg, Tiergartenstr. 17, 69121 Heidelberg, Germany
\institute{
University of Tyumen, Tyumen, Russia\\
\email{a.v.glazkova@utmn.ru}\\\and
Novosibirsk State University, Novosibirsk, Russia\\
\email{morozowdm@gmail.com}\\\and
The Institute for Information Transmission Problems (Kharkevich Institute), Moscow, Russia}
\maketitle              % typeset the header of the contribution
\begin{abstract}
Modern models for text generation show state-of-the-art results in many natural language processing tasks. In this work, we explore the effectiveness of abstractive text summarization models for keyphrase selection. A list of keyphrases is an important element of a text in databases and repositories of electronic documents. In our experiments, abstractive text summarization models fine-tuned for keyphrase generation show quite high results for a target text corpus. However, in most cases, the zero-shot performance on other corpora and domains is significantly lower. We investigate cross-domain limitations of abstractive text summarization models for keyphrase generation. We present an evaluation of the fine-tuned BART models for the keyphrase selection task across six benchmark corpora for keyphrase extraction including scientific texts from two domains and news texts. We explore the role of transfer learning between different domains to improve the BART model performance on small text corpora. Our experiments show that preliminary fine-tuning on out-of-domain corpora can be effective under conditions of a limited number of samples.

\end{abstract}
\section{Introduction}

The task of keyphrase generation aims at predicting a set of keyphrases summarizing the content of the source text. Keyphrases are often indexed in databases to improve the performance of information retrieval tools. Researchers select keyphrases for their papers to increase their visibility in the scientific community. Automatic selection of keyphrases for scholarly documents helps to analyze the current research trends, recommend papers, and identify potential peer reviewers~\cite{swaminathan2020preliminary}.

%\begin{table}[]
%\centering
%\begin{tabular}{|p{12cm}|}\hline
%\textbf{Abstract.} The study considers \underline{robust estimation} of \underline{linear regression parameters} by the regularization method, the \underline{pseudoinverse method}, and the \underline{Bayesian method} allowing for correlations and errors in the data. Regularizing algorithms are constructed and their relationship with \underline{pseudoinversion}, the \underline{Bayesian approach}, and \underline{BLUE} is investigated \\\hline
%\textbf{Keyphrases:} linear regression problems regularization, \underline{robust estimation}, \underline{linear regression parameters}, \underline{pseudoinverse method}, \underline{Bayesian method}, \underline{pseudoinversion}, \underline{Bayesian approach}, \underline{BLUE}, Bayes methods, estimation theory, probability, statistical analysis\\\hline
%\end{tabular}
%\end{table}

\begin{figure}[htp]
    \centering
    \includegraphics[width = 10 cm]{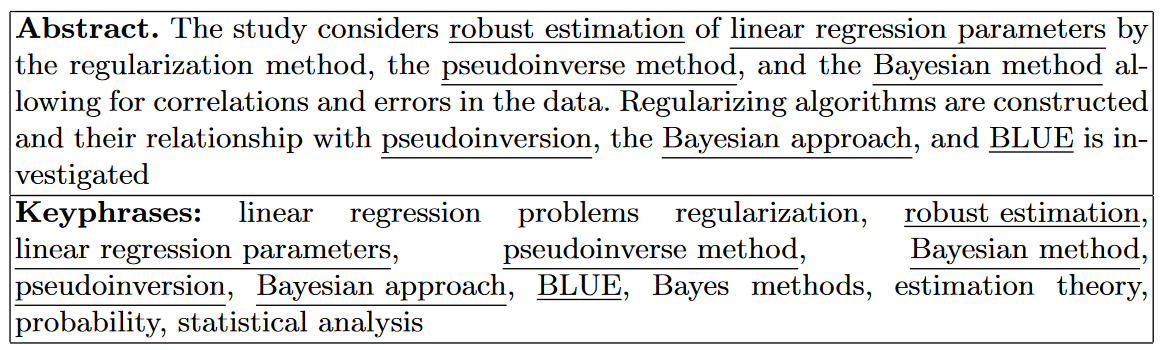}
    \caption{An example of a source text with the corresponding list of from the Inspec corpus~\cite{hulth2003improved}. The keyphrases that appear in the text are underlined.}
    \label{fig:example_keywords}
\end{figure}

Figure \ref{fig:example_keywords} demonstrates an example of a source text and its keyphrases. Some keyphrases are present in the source text while others are absent. Most unsupervised approaches for keyphrase selection have the purpose of keyphrase extraction, in other words, the ranking and selection of phrases that appear in the text. Recent generative approaches produce both keyphrases present in the text and those absent from it. These approaches utilize deep learning methods using the encoder-decoder architecture~\cite{ccano2019keyphrase,meng2017deep,ye2018semi} and various training techniques, such as incorporating a copying mechanism~\cite{wang2022automatic}, reinforcement learning~\cite{chan2019neural}, hierarchical decoding~\cite{chen2020exclusive}, and multitask learning~\cite{kulkarni2022learning}. 
Currently, the models for automatic text generation achieve high results in various natural language processing tasks. Since a list of keyphrases is some type of summary of a scientific text, pre-trained models for abstractive summarization appear to be effective for generating keyphrases as a sequence. In our previous work~\cite{glazkova2023multi,glazkova2023applying}, we explored the performance of some of these models for keyphrase generation. It was shown that BART~\cite{lewis2020bart} fine-tuned for generating lists of keyphrases on texts from the target domain showed competitive results as compared to several baselines. However, it can show lower performance on the texts from other corpora and domains similar to other fine-tuned models. Our goal is to evaluate whether we can transfer knowledge from the BART model that was fine-tuned to generate keyphrases for one domain to another ones. We seek to answer the following research questions:

\begin{itemize}
    \item \textbf{RQ1.} How effective is a text summarization model fine-tuned on one corpus or one domain for generating keyphrases from texts of other corpora or domains in zero-shot settings?
    \item \textbf{RQ2.} Can we improve the model performance by adding training examples from other corpora and domains?
    \item \textbf{RQ3.} With a small number of training examples, can the model perform as effectively as a model fine-tuned on larger corpora?
    \item \textbf{RQ4.} Can transfer learning improve the model performance using a varying size of training data?
\end{itemize}

The paper is organized as follows. Section 2 presents related works in the field. In Section 3, we describe the corpora. Section 4 contains a brief description of the models we used. Section 5 presents the experimental setup. The results are reported and discussed in Section 6. Section 7 concludes this paper.

\section{Related Work}

\subsection{Abstractive Text Summarization using Pre-trained Transformers}

Pre-trained language models show impressive results in many natural language processing (NLP) tasks. A pre-trained model is a saved network that is previously trained on a large dataset. This is a common and highly effective approach to deep learning on small datasets~\cite{dung2019autonomous}. Automatic text summarization is a relevant trend in NLP. A summary can be generated through extractive, as well as abstractive, methods. Abstractive methods are difficult to implement because they need a lot of natural language processing. However, abstractive models, such as BART~\cite{lewis2020bart}, PEGASUS~\cite{zhang2020pegasus}, and many others, allow us to generate novel samples by either rephrasing or using new words, instead of simply extracting the important sentences~\cite{gupta2019abstractive,tank2022text}.

Neural abstractive summarization based on pre-trained language models has been studied by many researchers and showed a high performance with the aid of large text corpora. In particular, abstractive summarization models were applied for generating summaries in news~\cite{bukhtiyarov2020advances,chen2021news,golovizninaautomatic,zolotareva2020abstractive}, scientific~\cite{cachola2020tldr,rubio2022hulat,wright2022generating}, sport~~\cite{malykh2021generating}, and financial domains~\cite{vaca2022extractive,zmandar2023comparative}. One of the main challenges related to neural abstractive summarization is that domain-shifting problems and overfitting could occur with a small number of samples for the target corpora~\cite{chen2021meta}. The use of additional texts for other corpora is not always successful since different corpora contain texts of different writing styles and forms. The annotation for abstractive summarization is costly. Therefore, exploring approaches to low-resource abstractive summarization is very relevant and attracts the attention of scientists.

\subsection{Keyword Selection}

Keyword selection approaches can be roughly divided into three categories: i) actual keyword extraction, ii) keyword assignment, and iii) keyphrase generation~~\cite{beliga2014keyword,ccano2019keyphrase}. The actual keyword extraction involves extracting words directly presented in the text. In keyword assignment, keywords are chosen from a predefined set of terms, while documents are classified into thematic categories according to their topics. Keyphrase generation aims to produce a set or string of keywords using recent advances in sequence-to-sequence applications of neural networks. In this work we focus on keyphrase generation but use some keyword extraction approaches as baselines. Keyphrase generation allows us to generate broad terms and keyphrases that are not presented in the source text in an explicit form.

To date, some scholars have examined neural models to generate multiple keyphrases as a sequence~\cite{ccano2019keyphrase,swaminathan2020preliminary}. Chowdhury et al.~\cite{chowdhury2022applying} demonstrated that fine-tuned BART shows competitive results in keyphrase generation compared with the existing extractive neural models. In~\cite{jiang2023generating,wu2022pre}, the authors experimented with controllable text generation for producing keyphrases. The authors of~\cite{kulkarni2022learning} proposed KeyBART, a new pre-training setup for the BART model that learns to generate keyphrases in their original order in the source document. Shen and Le~~\cite{shen2023enhanced} investigated the advantages of title attention and sequence code representing phrase order in a keyphrase sequence in improving Transformer-based keyphrase generation. 

The authors of~\cite{song2023survey} provided a comprehensive survey on recent advances in keyphrase selection from pre-trained language models. They emphasize that most existing keyphrase extraction datasets and studies are based on a few of the most common topics and lack datasets and research related to other domains. Therefore, transferring knowledge from one domain to another to build domain-specific keyphrase extraction models is one of the major challenges for keyphrase generation.

\section{Data}

\begin{table}[t]
\caption{A summary statistics for the corpora. The average number of tokens was obtained using NLTK~\cite{bird2006nltk}. The "$\pm$" sign is utilized to indicate a standard deviation. The abbreviations in this table are CS --- computer science, BM --- biomedical, A --- abstract, and T --- text (body).}
\centering
\begin{tabular}{|p{2.7cm}|p{1.1cm}|p{1.3cm}|p{1.1cm}|p{1.3cm}|p{1.3cm}|p{1.2cm}|p{1.2cm}|}\hline
Characteristic & \vertcell{2cm}{Krapivin-A} & \vertcell{2cm}{Krapivin-T} & \vertcell{2cm}{Inspec} & \vertcell{2cm}{PubMed} & \vertcell{2cm}{NamedKeys} & \vertcell{2cm}{DUC-2001} & \vertcell{2cm}{KPTimes} \\
\hline
Size & 2,294 & 2,293 & 2,000 &  1,320&  3,049&  308&  20,000\\\hline
Domain &  \multicolumn{3}{c|}{scientific, CS} &  \multicolumn{2}{c|}{scientific, BM}  &  \multicolumn{2}{c|}{\multirow{2}{*}{news}}  \\\cline{1-6}
Type of texts & \centering A & \centering T & \centering A & \centering T & \centering A &  \multicolumn{2}{c|}{} \\\hline
Avg. number of tokens & 169.06 $\pm$68.58 & 8597.63 $\pm$2411.77 & 127.35 $\pm$65.03 & 5270.97 $\pm$2690.67 & 274.67 $\pm$99.88 & 848.22 $\pm$563.41 &  733.78 $\pm$477.49\\\hline
Avg. number of sentences & 6.64 $\pm$2.69&343.95 $\pm$107.3 & 5.3 $\pm$2.73& 206.81 $\pm$127.11 &10.52 $\pm$3.66  & 34.74 $\pm$23.33  &  26.65 $\pm$22.19 \\\hline
Avg. keyphrases per text & \multicolumn{2}{c|}{5.34$\pm$2.77}  & 14.11 $\pm$6.41 &  5.4 $\pm$2.17& 14.15 $\pm$5.2 &  8.08 $\pm$1.87&  5.03 $\pm$1.88\\\hline
Absent keyphrases, \% &  51.3 $\pm$25.99& 18.04 $\pm$19.69 & 43.8 $\pm$17.83 & 13.52 $\pm$19.7 & 1.04 $\pm$5.83 &  2.45 $\pm$7.94 & 35.72 $\pm$29.22 \\\hline
Number of unique keyphrases &  \multicolumn{2}{c|}{8,703} & 19,066& 5,580&  20,804&  1,850&  21,126\\\hline
\end{tabular}
\label{summary}
\end{table}

\begin{table}[]
\caption{Top-10 common keyphrases for the corpora.}
\centering
\begin{tabular}{|p{2cm}|p{9.5cm}|}
\hline
Corpus & Keyphrases (keyphrase --- number of occurrences) \\ \hline
 Krapivin&  scheduling -- 36, performance evaluation -- 25, data mining -- 24, computational complexity -- 24, parallel algorithms -- 22, fault tolerance -- 22, approximation algorithms -- 22, model checking -- 21, distributed systems -- 21, preconditioning -- 20\\ \hline
 Inspec& internet -- 199, information resources -- 97, probability -- 70, computational complexity -- 69, optimisation -- 60, gender issues -- 49, matrix algebra -- 47, psychology -- 46, human factors -- 46, academic libraries -- 45 \\ \hline
 PubMed& children -- 24, breast cancer -- 21, epidemiology -- 20, internet -- 19, quality of life -- 19, preconception care -- 16, pregnancy -- 16, apoptosis -- 15, cancer -- 13, magnetic resonance imaging -- 13\\ \hline
 NamedKeys&  CI -- 245, OR -- 144, reactive oxygen species -- 142, ROS -- 134, confidence interval -- 131, nitric oxide -- 112, NO -- 111, HR -- 95, ER -- 84, oxidative stress -- 83\\ \hline
 Duc-2001& police brutality -- 12, mad cow disease -- 11, illegal aliens -- 10, Census Bureau -- 10, Ben Johnson -- 10, Clarence Thomas -- 10, investigation -- 9, firefighters -- 9, welfare reform -- 8, crash -- 8 \\ \hline
 KPTimes & U.S. -- 1,472, Donald Trump -- 1,274, China -- 1,122, terrorism -- 525, baseball -- 510, Russia -- 474, elections -- 435, Shinzo Abe -- 386, football -- 364, North Korea -- 350 \\ \hline
\end{tabular}
\label{popular}
\end{table}

\begin{table}[]
\caption{A comparison of corpora content proximity, evaluated as in~\cite{kilgarriff2001comparing}. The value of 1 indicates identical corpora. The higher the score, the greater the difference between corpora.}
\centering
\begin{tabular}{|p{2cm}|p{1.2cm}|p{1.3cm}|p{1.2cm}|p{1.2cm}|p{1.3cm}|p{1.2cm}|p{1.2cm}|}\hline
 & \vertcell{2cm}{Krapivin-A} & \vertcell{2cm}{Krapivin-T} & \vertcell{2cm}{Inspec} & \vertcell{2cm}{PubMed} & \vertcell{2cm}{NamedKeys} & \vertcell{2cm}{DUC-2001} & \vertcell{2cm}{KPTimes} \\
\hline
Krapivin-A & 1.00 & 46.24 & 51.63 & 141.92 & 214.31 & 295.17 & 267.85 \\\hline
Krapivin-T & 46.24 & 1.00 & 61.44 & 125.14 & 190.75 & 277.01 & 238.56 \\\hline
Inspec & 51.63 & 61.44 & 1.00 & 98.11 & 146.79 & 216.29 & 202.35 \\\hline
PubMed & 141.92 & 125.14 & 98.11 & 1.00 & 68.72 & 194.62 & 121.91 \\\hline
NamedKeys & 214.31 & 190.75 & 146.79 & 68.72 & 1.00 & 309.29 & 285.13 \\\hline
DUC-2001 & 295.17 & 277.01 & 216.29 & 194.62 & 309.29 & 1.00 & 162.96 \\\hline
KPTimes & 267.85 & 238.56 & 202.35 & 121.91 & 285.13 & 162.96 & 1.00 \\\hline
\end{tabular}
\label{sketchengine}
\end{table}

The experiments are carried out on six corpora for keyphrase selection: 
\begin{itemize}
    \item Krapivin~\cite{krapivin2009large} and Inspec~\cite{hulth2003improved} containing scientific texts from the computer science domain; 
    \item PubMed~\cite{schutz2008keyphrase} and NamedKeys~\cite{gero2019namedkeys}, which include scientific texts from the biomedical domain;
    \item DUC-2001~\cite{wan2008single} and KPTimes~\cite{gallina2019kptimes} consisting of news texts.
\end{itemize}  
The Krapivin corpus contains full papers divided into titles, abstracts, and bodies. In this work, we separately utilized the abstract and body of the paper to select keyphrases (Krapivin-A and Krapivin-T respectively). The original KPTimes corpus is composed of 279,923 article-keyphrase pairs. Here, we used only a test set of the original corpus containing 20,000 samples. Summary statistics for the corpora are presented in Table \ref{summary}. The most popular keyphrases are shown in Table \ref{popular}. A comparison of the contents of the corpora is given in Table \ref{sketchengine}.

\section{Models}

For keyphrase generation, we utilized BART-base~\cite{lewis2020bart}, a transformer-based denoising autoencoder for pre-training a seq2seq model. The model has 12 layers, 768 hidden units per layer, and a total of 139M parameters. BART was pre-trained by corrupting documents and then optimizing a reconstruction loss—the cross-entropy between the decoder’s output and the original document. We fine-tuned BART-base for six epochs with a maximum sequence length of 256 tokens. We utilized a standard cross-entropy loss and the AdamW optimizer \cite{loshchilov2018decoupled}. We used the source text as an input of the model and a list of keyphrases in a string format as an output. Keyphrases included in lists of keyphrases were separated with commas.

As baselines, we used the implementations of TopicRank~\cite{bougouin2013topicrank} and YAKE!~\cite{campos2020yake} from the PKE library~\cite{boudin2016pke} and KeyBART~\cite{kulkarni2022learning} that represents pre-trained BART-based architecture to produce a sequence of keyphrases pre-trained on the OAGKX dataset~\cite{ccano2020two}, which consists of 23 million scientific documents across multiple domains. 

\section{Experimental Setup}

We randomly split each corpus into a 70\% training set and a 30\% test set. For BART, we performed three runs for each model and then calculated the average results. Since TopicRank and YAKE! are unsupervised methods and they require a pre-defined number of keyphrases to select, we extracted 5, 10, and 15 keyphrases for each corpus and chose the best value for each metric. KeyBART was used in zero-shot settings. The models were evaluated in terms of the full-match F1-score (F1), ROUGE-1 (R1), ROUGE-L (RL)~\cite{lin2004rouge}, and BERTScore~(BS)~\cite{zhangbertscore}.

The full-match F1-score evaluates the number of exact matches between the original and generated sets of keyphrases. It is calculated as a harmonic mean of precision and recall.

The ROUGE-1 score calculates the number of matching unigrams between the model-generated text and the reference. The ROUGE-L score works in a similar way but measures the longest common subsequence. To measure ROUGE-1 and ROUGE-L, the keyphrases for each text were combined into a string with a comma as a separator. 

BERTScore utilizes the pre-trained contextual embeddings from BERT-based models and matches words in the source and generated
texts using cosine similarity. It has been shown that human judgment correlates with this metric on sentence-level and system-level evaluation. To calculate BERTScore, we use contextual embeddings from RoBERTa-large~\cite{liu2019roberta}, a modification of BERT that is pre-trained using dynamic masking.

\section{Results and Discussion}

To answer \textbf{RQ1} and \textbf{RQ2}, we fine-tuned BART on one corpus and applied it to the other corpora in zero-shot settings. Then we fine-tuned BART on mixed data. For this purpose, we evaluated four strategies: 

\begin{enumerate}
    \item \textit{Domain$_{eq}$}, fine-tuning the model on the texts of all corpora from one domain (for example, CS domain includes Krapivin-a, Krapivin-T, and Inspec), then testing on each corpus separately. In this strategy, we use an equal number of texts for each corpus. For example, if the size of training sets for Krapivin-A, Krapivin-T, and Inspec are 1,606, 1,605, and 1,400 respectively, we utilize 1,400 random texts from Krapivin-A and Krapivin-T and all texts from Inspec. The overall size of training data is 4,200. The texts from different corpora are mixed in random order.
    \item \textit{Domain$_{all}$}, the strategy is similar to the previous one but we use all texts from each corpus. In this case, the overall size of training data for the example above is 4,611, i.e. 1,606+1,605+1,400.
    \item \textit{Mix$_{eq}$}, fine-tuning the model on the texts of all corpora using an equal number of texts for each corpus, then testing it on each corpus separately. The texts from different corpora are mixed in random order.
    \item \textit{Mix$_{all}$}, the strategy is similar to \textit{Mix$_{eq}$} but we use all texts from each corpus.
\end{enumerate}

Table \ref{baselines} shows the performance of baselines on test sets. The best baseline results are underlined. The performance of different methods varies depending on the corpus. For example, KeyBART performs worse on the news domain since this model was pre-trained on scientific texts.

The BART results are presented in Table \ref{bart}. The results obtained for models fine-tuned on data containing the target corpus are highlighted in gray. Training data are italicized. The scores outperforming baselines are underlined. For mixed training data, we indicate the overall number of training examples in brackets and highlighted in bold the scores that exceed the results of the BART fine-tuned only on the target corpus. The best results among all models (Tables \ref{baselines} and \ref{bart}) are marked with an asterisk (*). Table \ref{appendix_table} in Appendix \ref{appendix} shows a standard deviation for three runs of BART.

\begin{table}[]
\caption{Baseline results, \%.}
\centering
\scriptsize
\begin{tabular}{|l|llll||llll||llll|}
\hline
\multicolumn{1}{|c|}{\multirow{2}{*}{Target}} & \multicolumn{1}{c|}{F1} & \multicolumn{1}{c|}{R1} & \multicolumn{1}{c|}{RL} & \multicolumn{1}{c|}{BS} & \multicolumn{1}{c|}{F1} & \multicolumn{1}{c|}{R1} & \multicolumn{1}{c|}{RL} & \multicolumn{1}{c|}{BS} & \multicolumn{1}{c|}{F1} & \multicolumn{1}{c|}{R1} & \multicolumn{1}{c|}{RL} & \multicolumn{1}{c|}{BS} \\ \cline{2-13} 
\multicolumn{1}{|c|}{} & \multicolumn{4}{c|}{\textit{KeyBART}} & \multicolumn{4}{c|}{\textit{YAKE!}} & \multicolumn{4}{c|}{\textit{TopicRank}} \\ \hline
Krapivin-A & \multicolumn{1}{l|}{\underline{8.58}} & \multicolumn{1}{l|}{\underline{23.34}} & \multicolumn{1}{l|}{\underline{19.81}} & \underline{88.26} & \multicolumn{1}{l|}{8.14} & \multicolumn{1}{l|}{20.75} & \multicolumn{1}{l|}{17.58} & 86.35 & \multicolumn{1}{l|}{6.89} & \multicolumn{1}{l|}{17.68} & \multicolumn{1}{l|}{14.86} & 87.94 \\ \hline
Krapivin-T & \multicolumn{1}{l|}{5.42} & \multicolumn{1}{l|}{\underline{16.61}} & \multicolumn{1}{l|}{\underline{14.67}} & 87.18 & \multicolumn{1}{l|}{\underline{7.09}} & \multicolumn{1}{l|}{16.43} & \multicolumn{1}{l|}{14.13} & 86.14 & \multicolumn{1}{l|}{5.89} & \multicolumn{1}{l|}{15.17} & \multicolumn{1}{l|}{12.48} & \underline{87.44} \\ \hline
Inspec & \multicolumn{1}{l|}{10.66} & \multicolumn{1}{l|}{29.09} & \multicolumn{1}{l|}{23.55} & 86.72 & \multicolumn{1}{l|}{13.84} & \multicolumn{1}{l|}{33.80} & \multicolumn{1}{l|}{\underline{27.00}} & 86.45 & \multicolumn{1}{l|}{\underline{16.32}} & \multicolumn{1}{l|}{\underline{35.68}} & \multicolumn{1}{l|}{25.01} & \underline{87.38} \\ \hline
PubMed & \multicolumn{1}{l|}{5.70} & \multicolumn{1}{l|}{13.41} & \multicolumn{1}{l|}{12.27} & 85.56 & \multicolumn{1}{l|}{\underline{13.35}} & \multicolumn{1}{l|}{\underline{20.00}} & \multicolumn{1}{l|}{\underline{17.28}} & 85.43 & \multicolumn{1}{l|}{11.11} & \multicolumn{1}{l|}{18.61} & \multicolumn{1}{l|}{15.41} & \underline{86.83} \\ \hline
NamedKeys & \multicolumn{1}{l|}{9.01} & \multicolumn{1}{l|}{21.11} & \multicolumn{1}{l|}{18.30} & 82.96 & \multicolumn{1}{l|}{\underline{20.80}} & \multicolumn{1}{l|}{\underline{30.62*}} & \multicolumn{1}{l|}{\underline{22.06*}} & \underline{84.80} & \multicolumn{1}{l|}{19.40} & \multicolumn{1}{l|}{30.55} & \multicolumn{1}{l|}{22.00} & \underline{84.80} \\ \hline
DUC-2001 & \multicolumn{1}{l|}{5.57} & \multicolumn{1}{l|}{11.54} & \multicolumn{1}{l|}{10.37} & 86.12 & \multicolumn{1}{l|}{13.58} & \multicolumn{1}{l|}{26.93} & \multicolumn{1}{l|}{22.16} & 85.63 & \multicolumn{1}{l|}{\underline{20.88*}} & \multicolumn{1}{l|}{\underline{30.91*}} & \multicolumn{1}{l|}{\underline{23.59*}} & \underline{88.51*} \\ \hline
KPTimes & \multicolumn{1}{l|}{4.50} & \multicolumn{1}{l|}{8.52} & \multicolumn{1}{l|}{7.87} & 83.95 & \multicolumn{1}{l|}{10.05} & \multicolumn{1}{l|}{\underline{18.92}} & \multicolumn{1}{l|}{\underline{16.18}} & \underline{84.83} & \multicolumn{1}{l|}{\underline{10.40}} & \multicolumn{1}{l|}{14.44} & \multicolumn{1}{l|}{12.74} & 86.24 \\ \hline
\end{tabular}
\label{baselines}
\end{table}

\begin{table}[t]
\caption{BART results, \%.}
\scriptsize
\centering
\addtolength{\tabcolsep}{-1pt}
\begin{tabular}{|l|llll||llll||llll|}
\hline
\multirow{3}{*}{Target} & \multicolumn{12}{c|}{Training data} \\ \cline{2-13} 
 & \multicolumn{1}{c|}{F1} & \multicolumn{1}{c|}{R1} & \multicolumn{1}{c|}{RL} & \multicolumn{1}{c|}{BS} & \multicolumn{1}{c|}{F1} & \multicolumn{1}{c|}{R1} & \multicolumn{1}{c|}{RL} & \multicolumn{1}{c|}{BS} & \multicolumn{1}{c|}{F1} & \multicolumn{1}{c|}{R1} & \multicolumn{1}{c|}{RL} & \multicolumn{1}{c|}{BS} \\ \cline{2-13} 
\multicolumn{1}{|c|}{} & \multicolumn{4}{c|}{\textit{Krapivin-A}} & \multicolumn{4}{c|}{\textit{Krapivin-T}} & \multicolumn{4}{c|}{\textit{Inspec}} \\ \hline
Krapivin-A & \multicolumn{1}{l|}{\cellcolor{gray!30}\underline{11.77}} & \multicolumn{1}{l|}{\cellcolor{gray!30}\underline{24.88*}} & \multicolumn{1}{l|}{\cellcolor{gray!30}\underline{21.46*}} & \cellcolor{gray!30}\underline{88.37} & \multicolumn{1}{l|}{8.38} & \multicolumn{1}{l|}{20.87} & \multicolumn{1}{l|}{17.98} & 86.86 & \multicolumn{1}{l|}{7.44} & \multicolumn{1}{l|}{21.61} & \multicolumn{1}{l|}{16.94} & 85.04 \\ \hline
Krapivin-T & \multicolumn{1}{l|}{\underline{8.25}} & \multicolumn{1}{l|}{\underline{18.05}} & \multicolumn{1}{l|}{\underline{16.05}} & 87.39 & \multicolumn{1}{l|}{\cellcolor{gray!30}\underline{7.17}} & \multicolumn{1}{l|}{\cellcolor{gray!30}\underline{17.25}} & \multicolumn{1}{l|}{\cellcolor{gray!30}\underline{15.65}} & \cellcolor{gray!30}87.02 & \multicolumn{1}{l|}{5.05} & \multicolumn{1}{l|}{15.51} & \multicolumn{1}{l|}{13.31} & 84.30 \\ \hline
Inspec & \multicolumn{1}{l|}{9.85} & \multicolumn{1}{l|}{24.01} & \multicolumn{1}{l|}{19.37} & 86.63 & \multicolumn{1}{l|}{6.66} & \multicolumn{1}{l|}{19.27} & \multicolumn{1}{l|}{15.97} & 85.76 & \multicolumn{1}{l|}{\cellcolor{gray!30}\underline{20.18*}} & \multicolumn{1}{l|}{\cellcolor{gray!30}\underline{42.11}} & \multicolumn{1}{l|}{\cellcolor{gray!30}\underline{35.09}} & \cellcolor{gray!30}\underline{88.42} \\ \hline
PubMed & \multicolumn{1}{l|}{7.97} & \multicolumn{1}{l|}{15.67} & \multicolumn{1}{l|}{13.67} & 85.74 & \multicolumn{1}{l|}{4.44} & \multicolumn{1}{l|}{11.25} & \multicolumn{1}{l|}{10.12} & 83.76 & \multicolumn{1}{l|}{6.61} & \multicolumn{1}{l|}{16.98} & \multicolumn{1}{l|}{14.01} & 83.62 \\ \hline
NamedKeys & \multicolumn{1}{l|}{9.34} & \multicolumn{1}{l|}{16.05} & \multicolumn{1}{l|}{13.63} & 83.44 & \multicolumn{1}{l|}{5.56} & \multicolumn{1}{l|}{11.87} & \multicolumn{1}{l|}{10.22} & 81.85 & \multicolumn{1}{l|}{8.89} & \multicolumn{1}{l|}{22.30} & \multicolumn{1}{l|}{16.68} & 82.58 \\ \hline
DUC-2001 & \multicolumn{1}{l|}{6.01} & \multicolumn{1}{l|}{12.84} & \multicolumn{1}{l|}{11.85} & 86.00 & \multicolumn{1}{l|}{3.19} & \multicolumn{1}{l|}{8.40} & \multicolumn{1}{l|}{7.80} & 84.34 & \multicolumn{1}{l|}{7.81} & \multicolumn{1}{l|}{17.56} & \multicolumn{1}{l|}{15.48} & 85.27 \\ \hline
KPTimes & \multicolumn{1}{l|}{3.84} & \multicolumn{1}{l|}{7.17} & \multicolumn{1}{l|}{6.63} & 83.64 & \multicolumn{1}{l|}{2.46} & \multicolumn{1}{l|}{5.35} & \multicolumn{1}{l|}{4.99} & 82.30 & \multicolumn{1}{l|}{6.67} & \multicolumn{1}{l|}{11.71} & \multicolumn{1}{l|}{10.25} & 83.52 \\ \hline \cline{2-13} 
\multicolumn{1}{|c|}{} & \multicolumn{4}{c|}{\textit{PubMed}} & \multicolumn{4}{c|}{\textit{NamedKeys}} & \multicolumn{4}{c|}{\textit{DUC-2001}} \\ \hline
Krapivin-A & \multicolumn{1}{l|}{6.92} & \multicolumn{1}{l|}{15.67} & \multicolumn{1}{l|}{13.88} & 86.81 & \multicolumn{1}{l|}{5.22} & \multicolumn{1}{l|}{11.88} & \multicolumn{1}{l|}{10.54} & 84.81 & \multicolumn{1}{l|}{4.50} & \multicolumn{1}{l|}{14.85} & \multicolumn{1}{l|}{13.05} & 85.34 \\ \hline
Krapivin-T & \multicolumn{1}{l|}{4.19} & \multicolumn{1}{l|}{11.35} & \multicolumn{1}{l|}{10.33} & 85.89 & \multicolumn{1}{l|}{2.97} & \multicolumn{1}{l|}{7.77} & \multicolumn{1}{l|}{7.21} & 83.38 & \multicolumn{1}{l|}{2.65} & \multicolumn{1}{l|}{9.93} & \multicolumn{1}{l|}{9.16} & 84.14 \\ \hline
Inspec & \multicolumn{1}{l|}{6.08} & \multicolumn{1}{l|}{16.33} & \multicolumn{1}{l|}{13.92} & 85.46 & \multicolumn{1}{l|}{4.48} & \multicolumn{1}{l|}{11.04} & \multicolumn{1}{l|}{9.51} & 82.97 & \multicolumn{1}{l|}{4.69} & \multicolumn{1}{l|}{16.22} & \multicolumn{1}{l|}{14.07} & 83.77 \\ \hline
PubMed & \multicolumn{1}{l|}{\cellcolor{gray!30}13.35} & \multicolumn{1}{l|}{\cellcolor{gray!30}\underline{20.96}} & \multicolumn{1}{l|}{\cellcolor{gray!30}\underline{18.69}} & \cellcolor{gray!30}\underline{86.89*} & \multicolumn{1}{l|}{10.62} & \multicolumn{1}{l|}{17.92} & \multicolumn{1}{l|}{15.43} & 84.55 & \multicolumn{1}{l|}{6.00} & \multicolumn{1}{l|}{16.14} & \multicolumn{1}{l|}{14.26} & 85.17 \\ \hline
NamedKeys & \multicolumn{1}{l|}{12.47} & \multicolumn{1}{l|}{20.31} & \multicolumn{1}{l|}{17.13} & 84.16 & \multicolumn{1}{l|}{\cellcolor{gray!30}20.11} & \multicolumn{1}{l|}{\cellcolor{gray!30}27.04} & \multicolumn{1}{l|}{\cellcolor{gray!30}\underline{22.53}} & \cellcolor{gray!30}\underline{85.31} & \multicolumn{1}{l|}{6.42} & \multicolumn{1}{l|}{15.64} & \multicolumn{1}{l|}{13.25} & 82.75 \\ \hline
DUC-2001 & \multicolumn{1}{l|}{5.23} & \multicolumn{1}{l|}{11.64} & \multicolumn{1}{l|}{10.81} & 85.79 & \multicolumn{1}{l|}{6.27} & \multicolumn{1}{l|}{12.65} & \multicolumn{1}{l|}{11.32} & 84.61 & \multicolumn{1}{l|}{\cellcolor{gray!30}11.78} & \multicolumn{1}{l|}{\cellcolor{gray!30}24.14} & \multicolumn{1}{l|}{\cellcolor{gray!30}20.65} & \cellcolor{gray!30}87.45 \\ \hline
KPTimes & \multicolumn{1}{l|}{6.38} & \multicolumn{1}{l|}{9.89} & \multicolumn{1}{l|}{9.06} & 84.46 & \multicolumn{1}{l|}{8.94} & \multicolumn{1}{l|}{12.24} & \multicolumn{1}{l|}{10.97} & 84.22 & \multicolumn{1}{l|}{2.80} & \multicolumn{1}{l|}{8.47} & \multicolumn{1}{l|}{7.78} & 83.44 \\ \hline

\multicolumn{1}{|c|}{} & \multicolumn{4}{c|}{\textit{KPTimes}} & \multicolumn{4}{c|}{\textit{CS$_{eq}$} (4,200)} & \multicolumn{4}{c|}{\textit{CS$_{all}$} (4,611)} \\ \hline
Krapivin-A & \multicolumn{1}{l|}{3.73} & \multicolumn{1}{l|}{7.58} & \multicolumn{1}{l|}{7.20} & 84.81 & \multicolumn{1}{l|}{\cellcolor{gray!30}\underline{12.04}} & \multicolumn{1}{l|}{\cellcolor{gray!30}\underline{24.35}} & \multicolumn{1}{l|}{\cellcolor{gray!30}\underline{21.14}} & \cellcolor{gray!30}88.24 & \multicolumn{1}{l|}{\cellcolor{gray!30}\underline{\textbf{12.08}}} & \multicolumn{1}{l|}{\cellcolor{gray!30}\underline{24.49}} & \multicolumn{1}{l|}{\cellcolor{gray!30}\underline{21.02}} & \cellcolor{gray!30}\underline{88.30} \\ \hline
Krapivin-T & \multicolumn{1}{l|}{2.35} & \multicolumn{1}{l|}{6.27} & \multicolumn{1}{l|}{6.00} & 84.55 & \multicolumn{1}{l|}{\cellcolor{gray!30}\underline{\textbf{8.32}}} & \multicolumn{1}{l|}{\cellcolor{gray!30}\underline{\textbf{18.41}}} & \multicolumn{1}{l|}{\cellcolor{gray!30}\underline{\textbf{16.38}}} & \cellcolor{gray!30}\textbf{87.42} & \multicolumn{1}{l|}{\cellcolor{gray!30}\underline{\textbf{8.36*}}} & \multicolumn{1}{l|}{\cellcolor{gray!30}\underline{\textbf{18.54*}}} & \multicolumn{1}{l|}{\cellcolor{gray!30}\underline{\textbf{16.50*}}} & \cellcolor{gray!30}\underline{\textbf{87.48}} \\ \hline
Inspec & \multicolumn{1}{l|}{3.79} & \multicolumn{1}{l|}{7.80} & \multicolumn{1}{l|}{7.16} & 83.07 & \multicolumn{1}{l|}{\cellcolor{gray!30}\underline{20.04}} & \multicolumn{1}{l|}{\cellcolor{gray!30}\underline{42.00}} & \multicolumn{1}{l|}{\cellcolor{gray!30}\underline{34.86}} & \cellcolor{gray!30}\underline{\textbf{88.47}} & \multicolumn{1}{l|}{\cellcolor{gray!30}\underline{20.01}} & \multicolumn{1}{l|}{\cellcolor{gray!30}\underline{42.09}} & \multicolumn{1}{l|}{\cellcolor{gray!30}\underline{34.88}} & \cellcolor{gray!30}\underline{\textbf{88.47}} \\ \hline
PubMed & \multicolumn{1}{l|}{8.22} & \multicolumn{1}{l|}{10.77} & \multicolumn{1}{l|}{10.02} & 85.02 & \multicolumn{1}{l|}{8.11} & \multicolumn{1}{l|}{16.66} & \multicolumn{1}{l|}{14.45} & 85.57 & \multicolumn{1}{l|}{8.20} & \multicolumn{1}{l|}{16.84} & \multicolumn{1}{l|}{14.55} & 85.68 \\ \hline
NamedKeys & \multicolumn{1}{l|}{6.31} & \multicolumn{1}{l|}{7.75} & \multicolumn{1}{l|}{7.16} & 81.76 & \multicolumn{1}{l|}{7.99} & \multicolumn{1}{l|}{16.60} & \multicolumn{1}{l|}{13.75} & 83.06 & \multicolumn{1}{l|}{7.82} & \multicolumn{1}{l|}{16.52} & \multicolumn{1}{l|}{13.62} & 83.01 \\ \hline
DUC-2001 & \multicolumn{1}{l|}{6.86} & \multicolumn{1}{l|}{14.64} & \multicolumn{1}{l|}{13.15} & 86.12 & \multicolumn{1}{l|}{4.54} & \multicolumn{1}{l|}{11.36} & \multicolumn{1}{l|}{10.58} & 85.76 & \multicolumn{1}{l|}{5.16} & \multicolumn{1}{l|}{12.09} & \multicolumn{1}{l|}{11.30} & 85.87 \\ \hline
KPTimes & \multicolumn{1}{l|}{\cellcolor{gray!30}\underline{30.97*}} & \multicolumn{1}{l|}{\cellcolor{gray!30}\underline{33.98*}} & \multicolumn{1}{l|}{\cellcolor{gray!30}\underline{28.92*}} & \cellcolor{gray!30}\underline{88.12*} & \multicolumn{1}{l|}{7.11} & \multicolumn{1}{l|}{10.81} & \multicolumn{1}{l|}{9.79} & 84.51 & \multicolumn{1}{l|}{7.43} & \multicolumn{1}{l|}{10.98} & \multicolumn{1}{l|}{10.05} & 84.58 \\ \hline

\multicolumn{1}{|c|}{} & \multicolumn{4}{c|}{\textit{BM$_{eq}$} (1,848)} & \multicolumn{4}{c|}{\textit{BM$_{all}$} (3,058)} & \multicolumn{4}{c|}{\textit{News$_{eq}$} (432)} \\ \hline
Krapivin-A & \multicolumn{1}{l|}{6.63} & \multicolumn{1}{l|}{14.89} & \multicolumn{1}{l|}{13.13} & 86.50 & \multicolumn{1}{l|}{6.30} & \multicolumn{1}{l|}{14.52} & \multicolumn{1}{l|}{12.70} & 86.11 & \multicolumn{1}{l|}{5.43} & \multicolumn{1}{l|}{13.61} & \multicolumn{1}{l|}{12.05} & 85.77 \\ \hline
Krapivin-T & \multicolumn{1}{l|}{4.63} & \multicolumn{1}{l|}{11.41} & \multicolumn{1}{l|}{10.52} & 85.66 & \multicolumn{1}{l|}{4.41} & \multicolumn{1}{l|}{11.20} & \multicolumn{1}{l|}{10.30} & 85.47 & \multicolumn{1}{l|}{3.34} & \multicolumn{1}{l|}{9.69} & \multicolumn{1}{l|}{8.96} & 84.89 \\ \hline
Inspec & \multicolumn{1}{l|}{6.11} & \multicolumn{1}{l|}{15.26} & \multicolumn{1}{l|}{13.10} & 85.10 & \multicolumn{1}{l|}{5.78} & \multicolumn{1}{l|}{14.64} & \multicolumn{1}{l|}{12.36} & 84.64 & \multicolumn{1}{l|}{5.91} & \multicolumn{1}{l|}{13.74} & \multicolumn{1}{l|}{12.09} & 83.70 \\ \hline
PubMed & \multicolumn{1}{l|}{\cellcolor{gray!30}13.29} & \multicolumn{1}{l|}{\cellcolor{gray!30}\underline{\textbf{21.19}}} & \multicolumn{1}{l|}{\cellcolor{gray!30}\underline{18.48}} & \cellcolor{gray!30}86.33 & \multicolumn{1}{l|}{\cellcolor{gray!30}13.24} & \multicolumn{1}{l|}{\cellcolor{gray!30}\underline{\textbf{21.39}}} & \multicolumn{1}{l|}{\cellcolor{gray!30}\underline{18.46}} & \cellcolor{gray!30}85.88 & \multicolumn{1}{l|}{7.39} & \multicolumn{1}{l|}{15.80} & \multicolumn{1}{l|}{14.34} & 85.78 \\ \hline
NamedKeys & \multicolumn{1}{l|}{\cellcolor{gray!30}18.44} & \multicolumn{1}{l|}{\cellcolor{gray!30}25.38} & \multicolumn{1}{l|}{\cellcolor{gray!30}21.21} & \cellcolor{gray!30}\underline{85.10} & \multicolumn{1}{l|}{\cellcolor{gray!30}\textbf{20.69}} & \multicolumn{1}{l|}{\cellcolor{gray!30}\textbf{27.77}} & \multicolumn{1}{l|}{\cellcolor{gray!30}\underline{\textbf{23.13}}} & \cellcolor{gray!30}\underline{\textbf{85.42}} & \multicolumn{1}{l|}{7.60} & \multicolumn{1}{l|}{15.03} & \multicolumn{1}{l|}{12.83} & 83.18 \\ \hline
DUC-2001 & \multicolumn{1}{l|}{6.37} & \multicolumn{1}{l|}{13.88} & \multicolumn{1}{l|}{12.48} & 85.80 & \multicolumn{1}{l|}{6.48} & \multicolumn{1}{l|}{13.53} & \multicolumn{1}{l|}{12.32} & 85.53 & \multicolumn{1}{l|}{\cellcolor{gray!30}\textbf{13.76}} & \multicolumn{1}{l|}{\cellcolor{gray!30}\textbf{24.64}} & \multicolumn{1}{l|}{\cellcolor{gray!30}\textbf{21.15}} & \cellcolor{gray!30}\textbf{87.83} \\ \hline
KPTimes & \multicolumn{1}{l|}{9.43} & \multicolumn{1}{l|}{12.50} & \multicolumn{1}{l|}{11.26} & 84.83 & \multicolumn{1}{l|}{8.85} & \multicolumn{1}{l|}{12.47} & \multicolumn{1}{l|}{11.21} & 84.60 & \multicolumn{1}{l|}{\cellcolor{gray!30}4.76} & \multicolumn{1}{l|}{\cellcolor{gray!30}9.34} & \multicolumn{1}{l|}{\cellcolor{gray!30}8.60} & \cellcolor{gray!30}84.56 \\ \hline

\multicolumn{1}{|c|}{} & \multicolumn{4}{c|}{\textit{News$_{all}$} (14,216)} & \multicolumn{4}{c|}{\textit{Mix$_{eq}$} (1,512)} & \multicolumn{4}{c|}{\textit{Mix$_{all}$} (21,885)} \\ \hline
Krapivin-A & \multicolumn{1}{l|}{4.68} & \multicolumn{1}{l|}{10.38} & \multicolumn{1}{l|}{9.50} & 85.61 & \multicolumn{1}{l|}{\cellcolor{gray!30}\underline{9.57}} & \multicolumn{1}{l|}{\cellcolor{gray!30}21.67} & \multicolumn{1}{l|}{\cellcolor{gray!30}18.61} & \cellcolor{gray!30}87.75 & \multicolumn{1}{l|}{\cellcolor{gray!30}\underline{\textbf{12.52*}}} & \multicolumn{1}{l|}{\cellcolor{gray!30}\underline{24.82}} & \multicolumn{1}{l|}{\cellcolor{gray!30}\underline{21.41}} & \cellcolor{gray!30}{\underline{\textbf{88.41*}}} \\ \hline
Krapivin-T & \multicolumn{1}{l|}{3.41} & \multicolumn{1}{l|}{9.61} & \multicolumn{1}{l|}{8.99} & 85.58 & \multicolumn{1}{l|}{\cellcolor{gray!30}6.10} & \multicolumn{1}{l|}{\cellcolor{gray!30}15.91} & \multicolumn{1}{l|}{\cellcolor{gray!30}14.27} & \cellcolor{gray!30}86.68 & \multicolumn{1}{l|}{\cellcolor{gray!30}\underline{\textbf{8.24}}} & \multicolumn{1}{l|}{\cellcolor{gray!30}\underline{\textbf{18.09}}} & \multicolumn{1}{l|}{\cellcolor{gray!30}\underline{\textbf{16.19}}} & \cellcolor{gray!30}\underline{\textbf{87.50*}} \\ \hline
Inspec & \multicolumn{1}{l|}{6.95} & \multicolumn{1}{l|}{15.77} & \multicolumn{1}{l|}{13.81} & 85.00 & \multicolumn{1}{l|}{\cellcolor{gray!30}13.47} & \multicolumn{1}{l|}{\cellcolor{gray!30}33.34} & \multicolumn{1}{l|}{\cellcolor{gray!30}26.65} & \cellcolor{gray!30}87.24 & \multicolumn{1}{l|}{\cellcolor{gray!30}\underline{20.00}} & \multicolumn{1}{l|}{\cellcolor{gray!30}\underline{\textbf{42.25*}}} & \multicolumn{1}{l|}{\cellcolor{gray!30}\underline{\textbf{35.10*}}} & \cellcolor{gray!30}\underline{\textbf{88.51*}} \\ \hline
PubMed & \multicolumn{1}{l|}{9.71} & \multicolumn{1}{l|}{13.77} & \multicolumn{1}{l|}{12.46} & 85.60 & \multicolumn{1}{l|}{\cellcolor{gray!30}13.19} & \multicolumn{1}{l|}{\cellcolor{gray!30}\underline{20.45}} & \multicolumn{1}{l|}{\cellcolor{gray!30}\underline{17.85}} & \cellcolor{gray!30}86.77 & \multicolumn{1}{l|}{\cellcolor{gray!30}\underline{\textbf{13.71*}}} & \multicolumn{1}{l|}{\cellcolor{gray!30}\underline{\textbf{21.89*}}} & \multicolumn{1}{l|}{\cellcolor{gray!30}\underline{\textbf{18.94*}}} & \cellcolor{gray!30}86.25 \\ \hline
NamedKeys & \multicolumn{1}{l|}{8.37} & \multicolumn{1}{l|}{11.75} & \multicolumn{1}{l|}{10.43} & 82.72 & \multicolumn{1}{l|}{\cellcolor{gray!30}13.40} & \multicolumn{1}{l|}{\cellcolor{gray!30}20.36} & \multicolumn{1}{l|}{\cellcolor{gray!30}17.20} & \cellcolor{gray!30}84.33 & \multicolumn{1}{l|}{\cellcolor{gray!30}\textbf{20.79}} & \multicolumn{1}{l|}{\cellcolor{gray!30}\textbf{27.93}} & \multicolumn{1}{l|}{\cellcolor{gray!30}\underline{\textbf{23.26*}}} & \cellcolor{gray!30}\underline{\textbf{85.53*}} \\ \hline
DUC-2001 & \multicolumn{1}{l|}{\cellcolor{gray!30}\textbf{12.76}} & \multicolumn{1}{l|}{\cellcolor{gray!30}23.62} & \multicolumn{1}{l|}{\cellcolor{gray!30}20.45} & \cellcolor{gray!30}\textbf{87.93} & \multicolumn{1}{l|}{\cellcolor{gray!30}\textbf{13.88}} & \multicolumn{1}{l|}{\cellcolor{gray!30}\textbf{24.15}} & \multicolumn{1}{l|}{\cellcolor{gray!30}\textbf{21.29}} & \cellcolor{gray!30}\textbf{87.96} & \multicolumn{1}{l|}{\cellcolor{gray!30}\textbf{13.31}} & \multicolumn{1}{l|}{\cellcolor{gray!30}\textbf{25.63}} & \multicolumn{1}{l|}{\cellcolor{gray!30}\textbf{22.60}} & \cellcolor{gray!30}\textbf{88.02} \\ \hline
KPTimes & \multicolumn{1}{l|}{\cellcolor{gray!30}\underline{30.53}} & \multicolumn{1}{l|}{\cellcolor{gray!30}\underline{33.78}} & \multicolumn{1}{l|}{\cellcolor{gray!30}\underline{28.79}} & \cellcolor{gray!30}\underline{88.10} & \multicolumn{1}{l|}{\cellcolor{gray!30}5.56} & \multicolumn{1}{l|}{\cellcolor{gray!30}9.96} & \multicolumn{1}{l|}{\cellcolor{gray!30}9.18} & \cellcolor{gray!30}84.80 & \multicolumn{1}{l|}{\cellcolor{gray!30}\underline{30.22}} & \multicolumn{1}{l|}{\cellcolor{gray!30}\underline{33.49}} & \multicolumn{1}{l|}{\cellcolor{gray!30}\underline{28.54}} & \cellcolor{gray!30}\underline{88.07} \\ \hline

\end{tabular}
\label{bart}
\end{table}
The BART fine-tuned on a target corpus outperforms baselines in many cases (Krapivin-A, Inspec, and KPTimes -- all metrics; Krapivin-T -- F1, R1, and RL; PubMed -- R1, RL, and BS; NamedKeys -- RL and BS). For DUC-2001, the results of BART are lower than the ones of unsupervised methods, which is probably due to the smaller size of this corpus. The out-of-corpus results are generally lower than the in-corpus ones. For example, when fine-tuning on Inspec (CS domain), the performance in terms of F1 is reduced by 37\% and 30\% for Krapivin-A and Krapivin-T respectively (both -- CS), by 51\% and 56\% for PubMed and NamedKeys (BM), and by 34\% and 78\% for DUC-2001 and KPTimes (news). The only exception is the model fine-tuned on Krapivin-A. For Krapivin-T, its results are higher than the in-corpus scores. Thus, fine-tuning on abstracts demonstrated higher scores than the fine-tuning on texts of the papers for the same corpus. The lengths of abstracts and texts were limited to the first 256 tokens due to restrictions on the length of the input sequence and resource limits.

\begin{figure}[]
    \centering
    \includegraphics[width = 12 cm]{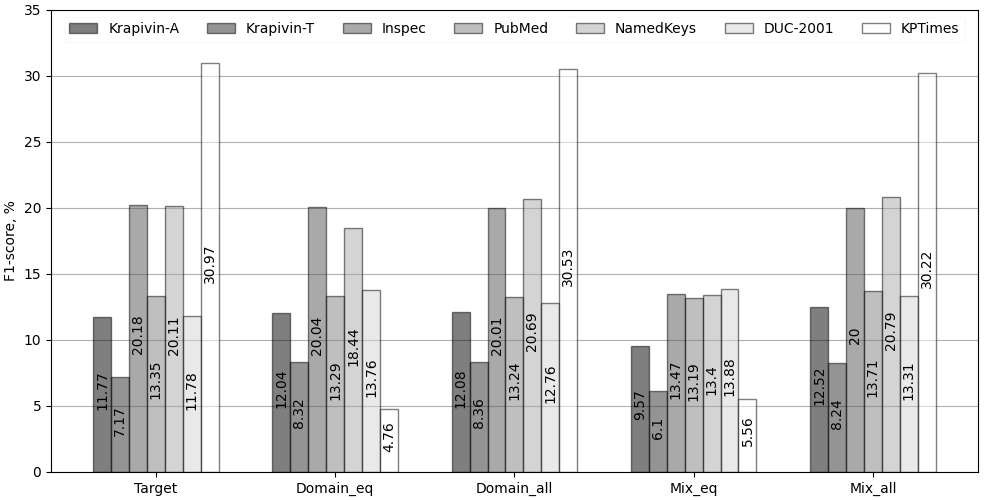}
    \caption{Adding training examples from other corpora and domains.}
    \label{fig:mix}
\end{figure}

Figure \ref{fig:mix} illustrates the effect of adding training examples from other corpora and domains in terms of F1. In our experiments, the effectiveness of the use of additional data varies depending on the characteristics of the corpus. For DUC-2001, which contains few training examples, the use of training examples from other corpora and domains increased the results for all strategies. In contrast, the highest result for KPTimes, which is the largest corpus in our experiments, is obtained using the only target training set. The use of the \textit{Domain$_{eq}$} and \textit{Mix$_{eq}$} strategies led to a sharp decrease in the size of the training set and the number of targeted examples and negatively affected the model performance. In general, the \textit{Mix$_{eq}$} strategy reduces the scores for all corpora except DUC-2001 due to a strong reduction in the amount of training data from the target corpus. \textit{Mix$_{all}$} generally improves the performance or at least does not lead to a strong degradation of results\footnote{This model is available at: \\\url{https://huggingface.co/beogradjanka/bart\_finetuned\_keyphrase\_extraction}}. This strategy showed the best results among all models for Krapivin-A (in terms of F1 and BS), Krapivin-T (BS), Inspec (R1, RL, and BS), PubMed (F1, R1, RL), and NamedKeys (RL, BS). Reducing the size of the dataset naturally leads to a decrease in training time. For instance, the training time is 53 minutes 59 seconds for \textit{Mix$_{all}$} (21,885 training examples) and 3 minutes 59 seconds for \textit{Mix$_{eq}$} (1,512 training examples). In this case, the training time decreases by about 20 times using the NVIDIA Tesla T4 GPU. 

To answer \textbf{RQ3}, we fine-tuned BART on a smaller number of training examples and evaluated the performance by increasing the size of the training data. Similarly to~\cite{miftahutdinov2020biomedical}, we used the following few-shot transfer procedure. We randomly sampled 50 texts from a target training set, fine-tuned the pre-trained model on this subset, and then tested it on a target test set. Next, we increased the sample size by 50 texts of the target training set and repeated the described procedure, doing so up to 1,000 texts or the end of the training set. We compared the results with the scores obtained using the full target corpus and the out-of-domain corpora mixed in equal proportions. For instance, for Krapivin-A, the mix of out-of-domain corpora includes PubMed, NamedKeys, DUC-2001, and KPTimes. To answer \textbf{RQ4}, we first fine-tuned BART on a mixture of out-of-domain corpora, and then fine-tuned the same model on the texts from the target corpus using the above strategy. We evaluated two options for two-stage fine-tuning. In the first case, we fine-tuned the model on out-of-domain data during half of the epochs (three epochs out of six) and then continued fine-tuning on the target data during the remaining three epochs. In the second case, we doubled the number of epochs and fine-tuned the model within six epochs on both out-of-domain and target data.

\begin{figure}
\begin{minipage}[h]{0.49\linewidth}
\center{\includegraphics[width=1\linewidth]{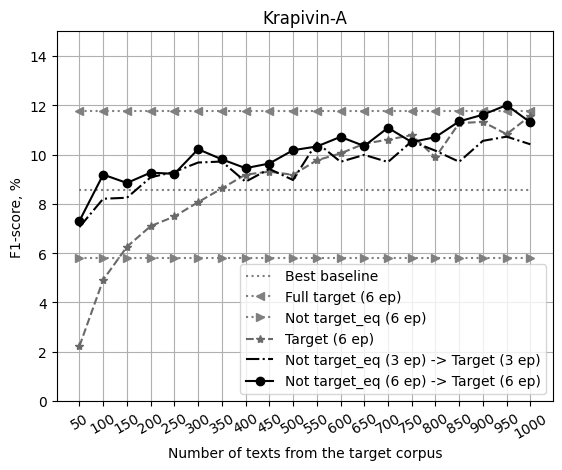}}  \\
\end{minipage}
\hfill
\begin{minipage}[h]{0.49\linewidth}
\center{\includegraphics[width=1\linewidth]{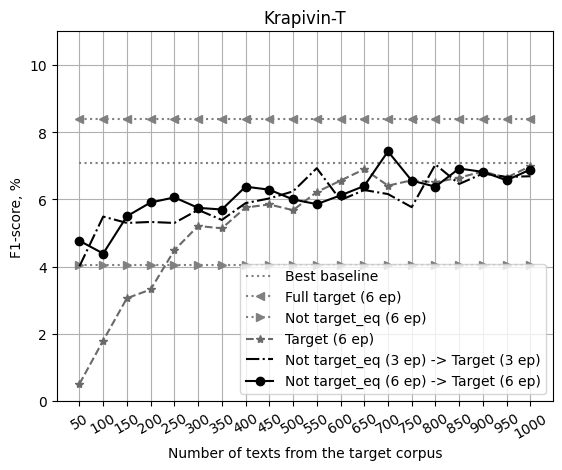}} \\
\end{minipage}
\vfill
\begin{minipage}[h]{0.49\linewidth}
\center{\includegraphics[width=1\linewidth]{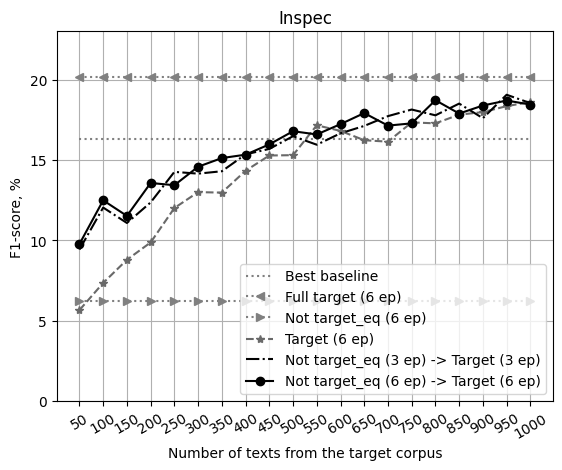}} \\
\end{minipage}
\hfill
\begin{minipage}[h]{0.49\linewidth}
\center{\includegraphics[width=1\linewidth]{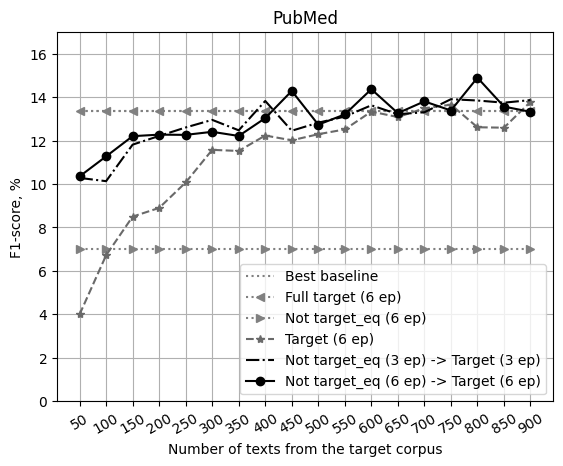}}  \\
\end{minipage}
\vfill
\begin{minipage}[h]{0.49\linewidth}
\center{\includegraphics[width=1\linewidth]{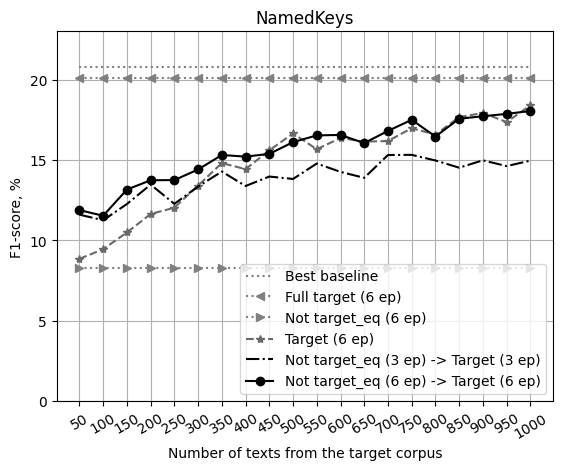}} \\
\end{minipage}
\hfill
\begin{minipage}[h]{0.49\linewidth}
\center{\includegraphics[width=1\linewidth]{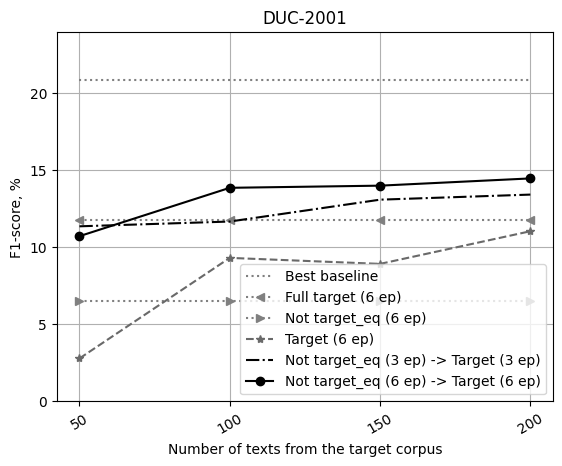}}  \\
\end{minipage}
\vfill
\begin{minipage}[h]{1\linewidth}
\center{\includegraphics[width=0.5\linewidth]{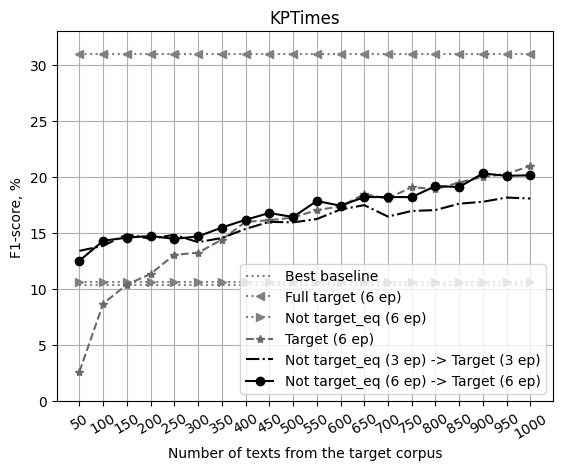}} \\
\end{minipage}
\caption{Performance of BART with and without preliminary fine-tuning on out-of-domain corpora using a varying size of training data.}
\label{ris:fsl}
\end{figure}

The results in terms of F1 are presented in Figure \ref{ris:fsl}. The figure uses the following conventions. \textit{Best baseline} -- the best baseline result for the dataset. \textit{Full target (6 ep)} -- fine-tuning on the full target corpus. \textit{Not target\_eq (6 ep)} -- fine-tuning on out-of-domain data. \textit{Target (6 ep)} -- fine-tuning on a part of the target corpus. \textit{Not target\_eq (3 ep) $\rightarrow$ Target (3 ep)} -- fine-tuning on out-of-domain data for three epochs, then fine-tuning on a part of a target corpus for three epochs. \textit{Not target\_eq (6 ep) $\rightarrow$ Target (6 ep)} -- fine-tuning on mixed out-of-domain data for six epochs, then fine-tuning on a part of the target corpus for six epochs. For PubMed, the best baseline result coincides with the line of Full target (6 ep).

The models with two-stage fine-tuning outperform the ones fine-tuned only on a target corpus on a small target sample size ($\approx$ up to 300 texts). Therefore, the use of out-of-domain corpora allows the use of fewer target data. For some corpora (Krapivin-A, PubMed, and DUC-2001), the models fine-tuned in a two-stage manner outperformed the ones fine-tuned on full target corpora. For Krapivin-A, the F1 outperforming the full target score was obtained using 59\% of the target training set. For PubMed and DUC-2001, it took us 43\% and 46\% respectively. For other corpora with a large full target size, we did not observe the results exceeding the F1 on a full target corpus during this experiment.

%\begin{figure}
%\begin{minipage}[h]{0.49\linewidth}
%\center{\includegraphics[width=1\linewidth]{mix.png}}  \\
%\end{minipage}
%\hfill
%\begin{minipage}[h]{0.49\linewidth}
%\center{\includegraphics[width=1\linewidth]{mix_r1.png}} \\
%\end{minipage}
%\vfill
%\begin{minipage}[h]{0.49\linewidth}
%\center{\includegraphics[width=1\linewidth]{mix_rl.png}} \\
%\end{minipage}
%\hfill
%\begin{minipage}[h]{0.49\linewidth}
%\center{\includegraphics[width=1\linewidth]{mix_bs.png}} \\
%\end{minipage}
%\caption{Performance of BART with and without preliminary fine-tuning on out-of-domain corpora using a varying size of training data.}
%\label{ris:adding}
%\end{figure}

\section{Conclusion}

We explored the robustness of the abstractive text summarization models fine-tuned for the task of keyphrase generation. Our experiments are based on BART, a transformer-based denoising autoencoder for pre-training a seq2seq model. We studied the cross-domain limitations of the BART fine-tuned for keyphrase generation across six corpora from three different domains. We also investigated the impact of preliminary out-of-domain fine-tuning to improve the performance of the models under conditions of a small amount of training data.

We found that preliminary fine-tuning on out-of-domain data improves the performance of the keyphrase generation in few-shot settings and allows the use of fewer target data.  Our findings add to a series of results concerning the effectiveness of a two-stage fine-tuning procedure where the transformer-based model is first fine-tuned on the source domain dataset before fine-tuning with the target domain dataset. For instance, similar studies conducted for text classification \cite{rietzler2020adapt,yadav2021forumbert} and named entity recognition \cite{miftahutdinov2020biomedical,syed2021menuner} have shown that the two-step training procedure can outperform the baseline models fine-tuned only on the target corpus. Our future research will focus on transfer learning from a high-resource language, for example, English, to other languages and to Russian in particular.
\newpage
\appendix
\section{Appendix}\label{appendix}

% \usepackage{multirow}
\begin{table}[]
\caption{The values of standard deviation for the BART results.}
\centering
\scriptsize
\begin{tabular}{|l|llll||llll||llll|}
\hline
\multicolumn{1}{|c|}{\multirow{2}{*}{Target corpus}} & \multicolumn{1}{c|}{F1} & \multicolumn{1}{c|}{R1} & \multicolumn{1}{c|}{RL} & \multicolumn{1}{c|}{BS} & \multicolumn{1}{c|}{F1} & \multicolumn{1}{c|}{R1} & \multicolumn{1}{c|}{RL} & \multicolumn{1}{c|}{BS} & \multicolumn{1}{c|}{F1} & \multicolumn{1}{c|}{R1} & \multicolumn{1}{c|}{RL} & \multicolumn{1}{c|}{BS} \\ \cline{2-13} 
\multicolumn{1}{|c|}{} & \multicolumn{4}{c|}{\textit{Krapivin-A}} & \multicolumn{4}{c|}{\textit{Krapivin-T}} & \multicolumn{4}{c|}{\textit{Inspec}} \\ \hline
Krapivin-A & \multicolumn{1}{l|}{0.36} & \multicolumn{1}{l|}{0.34} & \multicolumn{1}{l|}{0.25} & 0.05 & \multicolumn{1}{l|}{0.27} & \multicolumn{1}{l|}{0.54} & \multicolumn{1}{l|}{0.43} & 0.05 & \multicolumn{1}{l|}{0.40} & \multicolumn{1}{l|}{0.32} & \multicolumn{1}{l|}{0.48} & 0.11 \\ \hline
Krapivin-T & \multicolumn{1}{l|}{0.09} & \multicolumn{1}{l|}{0.16} & \multicolumn{1}{l|}{0.23} & 0.08 & \multicolumn{1}{l|}{0.18} & \multicolumn{1}{l|}{0.42} & \multicolumn{1}{l|}{0.35} & 0.05 & \multicolumn{1}{l|}{0.17} & \multicolumn{1}{l|}{0.16} & \multicolumn{1}{l|}{0.03} & 0.11 \\ \hline
Inspec & \multicolumn{1}{l|}{0.14} & \multicolumn{1}{l|}{0.26} & \multicolumn{1}{l|}{0.18} & 0.13 & \multicolumn{1}{l|}{0.25} & \multicolumn{1}{l|}{0.41} & \multicolumn{1}{l|}{0.24} & 0.07 & \multicolumn{1}{l|}{0.21} & \multicolumn{1}{l|}{0.29} & \multicolumn{1}{l|}{0.35} & 0.07 \\ \hline
PubMed & \multicolumn{1}{l|}{0.27} & \multicolumn{1}{l|}{0.10} & \multicolumn{1}{l|}{0.27} & 0.08 & \multicolumn{1}{l|}{0.30} & \multicolumn{1}{l|}{0.52} & \multicolumn{1}{l|}{0.38} & 0.17 & \multicolumn{1}{l|}{0.32} & \multicolumn{1}{l|}{0.30} & \multicolumn{1}{l|}{0.25} & 0.09 \\ \hline
NamedKeys & \multicolumn{1}{l|}{0.16} & \multicolumn{1}{l|}{0.75} & \multicolumn{1}{l|}{0.55} & 0.11 & \multicolumn{1}{l|}{0.40} & \multicolumn{1}{l|}{0.80} & \multicolumn{1}{l|}{0.57} & 0.21 & \multicolumn{1}{l|}{0.13} & \multicolumn{1}{l|}{0.26} & \multicolumn{1}{l|}{0.12} & 0.07 \\ \hline
DUC-2001 & \multicolumn{1}{l|}{0.92} & \multicolumn{1}{l|}{1.36} & \multicolumn{1}{l|}{1.21} & 0.17 & \multicolumn{1}{l|}{0.26} & \multicolumn{1}{l|}{0.47} & \multicolumn{1}{l|}{0.35} & 0.08 & \multicolumn{1}{l|}{4.15} & \multicolumn{1}{l|}{7.49} & \multicolumn{1}{l|}{6.35} & 0.84 \\ \hline
KPTimes & \multicolumn{1}{l|}{0.17} & \multicolumn{1}{l|}{0.23} & \multicolumn{1}{l|}{0.21} & 0.08 & \multicolumn{1}{l|}{0.25} & \multicolumn{1}{l|}{0.50} & \multicolumn{1}{l|}{0.46} & 0.16 & \multicolumn{1}{l|}{3.43} & \multicolumn{1}{l|}{5.21} & \multicolumn{1}{l|}{4.29} & 0.98 \\ \hline

\multicolumn{1}{|c|}{} & \multicolumn{4}{c|}{\textit{PubMed}} & \multicolumn{4}{c|}{\textit{NamedKeys}} & \multicolumn{4}{c|}{\textit{DUC-2001}} \\ \hline
Krapivin-A & \multicolumn{1}{l|}{0.38} & \multicolumn{1}{l|}{0.48} & \multicolumn{1}{l|}{0.43} & 0.18 & \multicolumn{1}{l|}{0.12} & \multicolumn{1}{l|}{0.05} & \multicolumn{1}{l|}{0.26} & 0.03 & \multicolumn{1}{l|}{0.23} & \multicolumn{1}{l|}{0.22} & \multicolumn{1}{l|}{0.33} & 0.08 \\ \hline
Krapivin-T & \multicolumn{1}{l|}{0.11} & \multicolumn{1}{l|}{0.17} & \multicolumn{1}{l|}{0.18} & 0.20 & \multicolumn{1}{l|}{0.27} & \multicolumn{1}{l|}{0.29} & \multicolumn{1}{l|}{0.27} & 0.20 & \multicolumn{1}{l|}{0.06} & \multicolumn{1}{l|}{0.15} & \multicolumn{1}{l|}{0.15} & 0.06 \\ \hline
Inspec & \multicolumn{1}{l|}{0.31} & \multicolumn{1}{l|}{0.37} & \multicolumn{1}{l|}{0.20} & 0.22 & \multicolumn{1}{l|}{0.21} & \multicolumn{1}{l|}{0.06} & \multicolumn{1}{l|}{0.16} & 0.11 & \multicolumn{1}{l|}{0.48} & \multicolumn{1}{l|}{0.88} & \multicolumn{1}{l|}{0.56} & 0.22 \\ \hline
PubMed & \multicolumn{1}{l|}{0.25} & \multicolumn{1}{l|}{0.43} & \multicolumn{1}{l|}{0.51} & 0.06 & \multicolumn{1}{l|}{0.39} & \multicolumn{1}{l|}{0.57} & \multicolumn{1}{l|}{0.54} & 0.22 & \multicolumn{1}{l|}{0.77} & \multicolumn{1}{l|}{0.91} & \multicolumn{1}{l|}{0.64} & 0.16 \\ \hline
NamedKeys & \multicolumn{1}{l|}{0.32} & \multicolumn{1}{l|}{0.55} & \multicolumn{1}{l|}{0.40} & 0.07 & \multicolumn{1}{l|}{0.84} & \multicolumn{1}{l|}{1.03} & \multicolumn{1}{l|}{0.77} & 0.23 & \multicolumn{1}{l|}{0.15} & \multicolumn{1}{l|}{0.16} & \multicolumn{1}{l|}{0.04} & 0.05 \\ \hline
DUC-2001 & \multicolumn{1}{l|}{0.24} & \multicolumn{1}{l|}{0.80} & \multicolumn{1}{l|}{0.75} & 0.41 & \multicolumn{1}{l|}{1.55} & \multicolumn{1}{l|}{1.44} & \multicolumn{1}{l|}{0.84} & 0.11 & \multicolumn{1}{l|}{0.52} & \multicolumn{1}{l|}{0.99} & \multicolumn{1}{l|}{0.48} & 0.16 \\ \hline
KPTimes & \multicolumn{1}{l|}{0.10} & \multicolumn{1}{l|}{0.13} & \multicolumn{1}{l|}{0.10} & 0.02 & \multicolumn{1}{l|}{0.25} & \multicolumn{1}{l|}{0.28} & \multicolumn{1}{l|}{0.28} & 0.14 & \multicolumn{1}{l|}{0.14} & \multicolumn{1}{l|}{0.05} & \multicolumn{1}{l|}{0.07} & 0.12 \\ \hline

\multicolumn{1}{|c|}{} & \multicolumn{4}{c|}{\textit{KPTimes}} & \multicolumn{4}{c|}{\textit{CS$_{eq}$}} & \multicolumn{4}{c|}{\textit{CS$_{all}$}} \\ \hline
Krapivin-A & \multicolumn{1}{l|}{0.18} & \multicolumn{1}{l|}{0.28} & \multicolumn{1}{l|}{0.25} & 0.14 & \multicolumn{1}{l|}{0.27} & \multicolumn{1}{l|}{0.80} & \multicolumn{1}{l|}{0.60} & 0.06 & \multicolumn{1}{l|}{0.04} & \multicolumn{1}{l|}{0.14} & \multicolumn{1}{l|}{0.13} & 0.05 \\ \hline
Krapivin-T & \multicolumn{1}{l|}{0.21} & \multicolumn{1}{l|}{0.40} & \multicolumn{1}{l|}{0.35} & 0.10 & \multicolumn{1}{l|}{0.28} & \multicolumn{1}{l|}{0.43} & \multicolumn{1}{l|}{0.31} & 0.02 & \multicolumn{1}{l|}{0.22} & \multicolumn{1}{l|}{0.12} & \multicolumn{1}{l|}{0.24} & 0.11 \\ \hline
Inspec & \multicolumn{1}{l|}{0.28} & \multicolumn{1}{l|}{0.30} & \multicolumn{1}{l|}{0.26} & 0.16 & \multicolumn{1}{l|}{0.60} & \multicolumn{1}{l|}{0.46} & \multicolumn{1}{l|}{0.46} & 0.08 & \multicolumn{1}{l|}{0.44} & \multicolumn{1}{l|}{0.41} & \multicolumn{1}{l|}{0.45} & 0.02 \\ \hline
PubMed & \multicolumn{1}{l|}{0.59} & \multicolumn{1}{l|}{0.71} & \multicolumn{1}{l|}{0.70} & 0.17 & \multicolumn{1}{l|}{0.48} & \multicolumn{1}{l|}{1.00} & \multicolumn{1}{l|}{0.99} & 0.24 & \multicolumn{1}{l|}{0.27} & \multicolumn{1}{l|}{0.26} & \multicolumn{1}{l|}{0.45} & 0.09 \\ \hline
NamedKeys & \multicolumn{1}{l|}{0.12} & \multicolumn{1}{l|}{0.28} & \multicolumn{1}{l|}{0.26} & 0.09 & \multicolumn{1}{l|}{0.23} & \multicolumn{1}{l|}{0.36} & \multicolumn{1}{l|}{0.28} & 0.09 & \multicolumn{1}{l|}{0.12} & \multicolumn{1}{l|}{0.54} & \multicolumn{1}{l|}{0.39} & 0.10 \\ \hline
DUC-2001 & \multicolumn{1}{l|}{0.96} & \multicolumn{1}{l|}{0.65} & \multicolumn{1}{l|}{0.70} & 0.26 & \multicolumn{1}{l|}{0.65} & \multicolumn{1}{l|}{1.32} & \multicolumn{1}{l|}{0.96} & 0.07 & \multicolumn{1}{l|}{0.43} & \multicolumn{1}{l|}{1.33} & \multicolumn{1}{l|}{0.98} & 0.28 \\ \hline
KPTimes & \multicolumn{1}{l|}{0.22} & \multicolumn{1}{l|}{0.19} & \multicolumn{1}{l|}{0.21} & 0.04 & \multicolumn{1}{l|}{0.56} & \multicolumn{1}{l|}{0.70} & \multicolumn{1}{l|}{0.54} & 0.12 & \multicolumn{1}{l|}{0.25} & \multicolumn{1}{l|}{0.09} & \multicolumn{1}{l|}{0.02} & 0.05 \\ \hline

\multicolumn{1}{|c|}{} & \multicolumn{4}{c|}{\textit{BM$_{eq}$}} & \multicolumn{4}{c|}{\textit{BM$_{all}$}} & \multicolumn{4}{c|}{\textit{News$_{eq}$}} \\ \hline
Krapivin-A & \multicolumn{1}{l|}{0.27} & \multicolumn{1}{l|}{0.30} & \multicolumn{1}{l|}{0.28} & 0.18 & \multicolumn{1}{l|}{0.14} & \multicolumn{1}{l|}{0.31} & \multicolumn{1}{l|}{0.32} & 0.13 & \multicolumn{1}{l|}{0.28} & \multicolumn{1}{l|}{0.63} & \multicolumn{1}{l|}{0.55} & 0.13 \\ \hline
Krapivin-T & \multicolumn{1}{l|}{0.11} & \multicolumn{1}{l|}{0.06} & \multicolumn{1}{l|}{0.23} & 0.22 & \multicolumn{1}{l|}{0.08} & \multicolumn{1}{l|}{0.16} & \multicolumn{1}{l|}{0.28} & 0.17 & \multicolumn{1}{l|}{0.08} & \multicolumn{1}{l|}{0.26} & \multicolumn{1}{l|}{0.28} & 0.09 \\ \hline
Inspec & \multicolumn{1}{l|}{0.09} & \multicolumn{1}{l|}{0.69} & \multicolumn{1}{l|}{0.53} & 0.14 & \multicolumn{1}{l|}{0.25} & \multicolumn{1}{l|}{0.49} & \multicolumn{1}{l|}{0.46} & 0.23 & \multicolumn{1}{l|}{0.43} & \multicolumn{1}{l|}{1.24} & \multicolumn{1}{l|}{0.93} & 0.46 \\ \hline
PubMed & \multicolumn{1}{l|}{0.83} & \multicolumn{1}{l|}{0.38} & \multicolumn{1}{l|}{0.41} & 0.10 & \multicolumn{1}{l|}{0.48} & \multicolumn{1}{l|}{0.06} & \multicolumn{1}{l|}{0.33} & 0.10 & \multicolumn{1}{l|}{0.43} & \multicolumn{1}{l|}{0.92} & \multicolumn{1}{l|}{0.75} & 0.07 \\ \hline
NamedKeys & \multicolumn{1}{l|}{0.52} & \multicolumn{1}{l|}{0.59} & \multicolumn{1}{l|}{0.37} & 0.13 & \multicolumn{1}{l|}{0.14} & \multicolumn{1}{l|}{0.01} & \multicolumn{1}{l|}{0.22} & 0.05 & \multicolumn{1}{l|}{0.22} & \multicolumn{1}{l|}{0.45} & \multicolumn{1}{l|}{0.39} & 0.07 \\ \hline
DUC-2001 & \multicolumn{1}{l|}{0.69} & \multicolumn{1}{l|}{2.30} & \multicolumn{1}{l|}{2.30} & 0.52 & \multicolumn{1}{l|}{0.07} & \multicolumn{1}{l|}{0.62} & \multicolumn{1}{l|}{0.52} & 0.17 & \multicolumn{1}{l|}{0.41} & \multicolumn{1}{l|}{0.51} & \multicolumn{1}{l|}{1.42} & 0.19 \\ \hline
KPTimes & \multicolumn{1}{l|}{0.28} & \multicolumn{1}{l|}{0.48} & \multicolumn{1}{l|}{0.35} & 0.01 & \multicolumn{1}{l|}{0.29} & \multicolumn{1}{l|}{0.11} & \multicolumn{1}{l|}{0.15} & 0.06 & \multicolumn{1}{l|}{0.05} & \multicolumn{1}{l|}{0.11} & \multicolumn{1}{l|}{0.14} & 0.05 \\ \hline

\multicolumn{1}{|c|}{} & \multicolumn{4}{c|}{\textit{News$_{all}$}} & \multicolumn{4}{c|}{\textit{Mix$_{eq}$}} & \multicolumn{4}{c|}{\textit{Mix$_{all}$}} \\ \hline
Krapivin-A & \multicolumn{1}{l|}{0.56} & \multicolumn{1}{l|}{0.90} & \multicolumn{1}{l|}{0.72} & 0.25 & \multicolumn{1}{l|}{0.50} & \multicolumn{1}{l|}{0.51} & \multicolumn{1}{l|}{0.68} & 0.09 & \multicolumn{1}{l|}{0.50} & \multicolumn{1}{l|}{0.52} & \multicolumn{1}{l|}{0.23} & 0.07 \\ \hline
Krapivin-T & \multicolumn{1}{l|}{0.12} & \multicolumn{1}{l|}{0.58} & \multicolumn{1}{l|}{0.47} & 0.19 & \multicolumn{1}{l|}{0.36} & \multicolumn{1}{l|}{0.34} & \multicolumn{1}{l|}{0.28} & 0.21 & \multicolumn{1}{l|}{0.02} & \multicolumn{1}{l|}{0.57} & \multicolumn{1}{l|}{0.42} & 0.10 \\ \hline
Inspec & \multicolumn{1}{l|}{0.53} & \multicolumn{1}{l|}{1.85} & \multicolumn{1}{l|}{1.28} & 0.33 & \multicolumn{1}{l|}{0.30} & \multicolumn{1}{l|}{0.20} & \multicolumn{1}{l|}{0.31} & 0.10 & \multicolumn{1}{l|}{0.33} & \multicolumn{1}{l|}{0.22} & \multicolumn{1}{l|}{0.46} & 0.02 \\ \hline
PubMed & \multicolumn{1}{l|}{0.15} & \multicolumn{1}{l|}{0.22} & \multicolumn{1}{l|}{0.22} & 0.10 & \multicolumn{1}{l|}{0.84} & \multicolumn{1}{l|}{0.20} & \multicolumn{1}{l|}{0.11} & 0.02 & \multicolumn{1}{l|}{0.62} & \multicolumn{1}{l|}{0.35} & \multicolumn{1}{l|}{0.42} & 0.07 \\ \hline
NamedKeys & \multicolumn{1}{l|}{0.55} & \multicolumn{1}{l|}{0.73} & \multicolumn{1}{l|}{0.52} & 0.21 & \multicolumn{1}{l|}{0.23} & \multicolumn{1}{l|}{0.38} & \multicolumn{1}{l|}{0.33} & 0.06 & \multicolumn{1}{l|}{0.21} & \multicolumn{1}{l|}{0.10} & \multicolumn{1}{l|}{0.02} & 0.08 \\ \hline
DUC-2001 & \multicolumn{1}{l|}{0.74} & \multicolumn{1}{l|}{0.27} & \multicolumn{1}{l|}{0.42} & 0.19 & \multicolumn{1}{l|}{0.54} & \multicolumn{1}{l|}{0.31} & \multicolumn{1}{l|}{0.20} & 0.08 & \multicolumn{1}{l|}{0.24} & \multicolumn{1}{l|}{0.73} & \multicolumn{1}{l|}{1.09} & 0.09 \\ \hline
KPTimes & \multicolumn{1}{l|}{0.12} & \multicolumn{1}{l|}{0.13} & \multicolumn{1}{l|}{0.17} & 0.01 & \multicolumn{1}{l|}{0.16} & \multicolumn{1}{l|}{0.13} & \multicolumn{1}{l|}{0.11} & 0.01 & \multicolumn{1}{l|}{0.28} & \multicolumn{1}{l|}{0.28} & \multicolumn{1}{l|}{0.17} & 0.02 \\ \hline

\end{tabular}
\label{appendix_table}
\end{table}

\bibliographystyle{samplepaper}
\bibliography{samplepaper}

\begin{thebibliography}{10}
\providecommand{\url}[1]{\texttt{#1}}
\providecommand{\urlprefix}{URL }
\providecommand{\doi}[1]{https://doi.org/#1}

\bibitem{beliga2014keyword}
Beliga, S.: Keyword extraction: a review of methods and approaches. University
  of Rijeka, Department of Informatics, Rijeka  \textbf{1}(9) (2014)

\bibitem{bird2006nltk}
Bird, S.: {NLTK}: the natural language toolkit. In: Proceedings of the
  COLING/ACL 2006 Interactive Presentation Sessions. pp. 69--72 (2006).
  \doi{10.3115/1225403.1225421}

\bibitem{boudin2016pke}
Boudin, F.: {PKE}: an open source python-based keyphrase extraction toolkit.
  In: Proceedings of COLING 2016, the 26th international conference on
  computational linguistics: system demonstrations. pp. 69--73 (2016)

\bibitem{bougouin2013topicrank}
Bougouin, A., Boudin, F., Daille, B.: Topic{R}ank: Graph-based topic ranking
  for keyphrase extraction. In: International joint conference on natural
  language processing (IJCNLP). pp. 543--551 (2013)

\bibitem{bukhtiyarov2020advances}
Bukhtiyarov, A., Gusev, I.: Advances of transformer-based models for news
  headline generation. In: Artificial Intelligence and Natural Language: 9th
  Conference, AINL 2020, Helsinki, Finland, October 7--9, 2020, Proceedings 9.
  pp. 54--61. Springer (2020). \doi{10.1007/978-3-030-59082-6_4}

\bibitem{cachola2020tldr}
Cachola, I., Lo, K., Cohan, A., Weld, D.S.: {TLDR}: Extreme summarization of
  scientific documents. In: Findings of the Association for Computational
  Linguistics: EMNLP 2020. pp. 4766--4777 (2020).
  \doi{10.18653/v1/2020.findings-emnlp.428}

\bibitem{campos2020yake}
Campos, R., Mangaravite, V., Pasquali, A., Jorge, A., Nunes, C., Jatowt, A.:
  {YAKE}! keyword extraction from single documents using multiple local
  features. Information Sciences  \textbf{509},  257--289 (2020).
  \doi{10.1016/j.ins.2019.09.013}

\bibitem{ccano2019keyphrase}
{\c{C}}ano, E., Bojar, O.: Keyphrase generation: A text summarization struggle.
  In: Proceedings of the 2019 Conference of the North American Chapter of the
  Association for Computational Linguistics: Human Language Technologies,
  Volume 1 (Long and Short Papers). pp. 666--672 (2019).
  \doi{10.18653/v1/n19-1070}

\bibitem{ccano2020two}
{\c{C}}ano, E., Bojar, O.: Two huge title and keyword generation corpora of
  research articles. In: Proceedings of the 12th Language Resources and
  Evaluation Conference. pp. 6663--6671 (2020)

\bibitem{chan2019neural}
Chan, H.P., Chen, W., Wang, L., King, I.: Neural keyphrase generation via
  reinforcement learning with adaptive rewards. In: Proceedings of the 57th
  Annual Meeting of the Association for Computational Linguistics. pp.
  2163--2174 (2019). \doi{10.18653/v1/p19-1208}

\bibitem{chen2020exclusive}
Chen, W., Chan, H.P., Li, P., King, I.: Exclusive hierarchical decoding for
  deep keyphrase generation. In: Proceedings of the 58th Annual Meeting of the
  Association for Computational Linguistics. pp. 1095--1105 (2020).
  \doi{10.18653/v1/2020.acl-main.103}

\bibitem{chen2021meta}
Chen, Y.S., Shuai, H.H.: Meta-transfer learning for low-resource abstractive
  summarization. In: Proceedings of the AAAI Conference on Artificial
  Intelligence. vol.~35, pp. 12692--12700 (2021).
  \doi{10.1609/aaai.v35i14.17503}

\bibitem{chen2021news}
Chen, Y., Song, Q.: News text summarization method based on {BART-TextRank}
  model. In: 2021 IEEE 5th Advanced Information Technology, Electronic and
  Automation Control Conference (IAEAC). pp. 2005--2010. IEEE (2021).
  \doi{10.1109/iaeac50856.2021.9390683}

\bibitem{chowdhury2022applying}
Chowdhury, M.F.M., Rossiello, G., Glass, M., Mihindukulasooriya, N., Gliozzo,
  A.: Applying a generic sequence-to-sequence model for simple and effective
  keyphrase generation. arXiv preprint arXiv:2201.05302  (2022).
  \doi{10.48550/arXiv.2201.05302}

\bibitem{dung2019autonomous}
Dung, C.V., et~al.: Autonomous concrete crack detection using deep fully
  convolutional neural network. Automation in Construction  \textbf{99},
  52--58 (2019). \doi{10.1016/j.autcon.2018.11.028}

\bibitem{gallina2019kptimes}
Gallina, Y., Boudin, F., Daille, B.: {KPTimes}: A large-scale dataset for
  keyphrase generation on news documents. In: Proceedings of the 12th
  International Conference on Natural Language Generation. pp. 130--135 (2019).
  \doi{10.18653/v1/w19-8617}

\bibitem{gero2019namedkeys}
Gero, Z., Ho, J.C.: Named{K}eys: Unsupervised keyphrase extraction for
  biomedical documents. In: Proceedings of the 10th ACM International
  Conference on Bioinformatics, Computational Biology and Health Informatics.
  pp. 328--337 (2019). \doi{10.1145/3307339.3342147}

\bibitem{glazkova2023multi}
Glazkova, A., Morozov, D.: Multi-task fine-tuning for generating keyphrases in
  a scientific domain. In: 2023 IX International Conference on Information
  Technology and Nanotechnology (ITNT). pp.~1--5. IEEE (2023).
  \doi{10.1109/ITNT57377.2023.10139061}

\bibitem{glazkova2023applying}
Glazkova, A., Morozov, D.: Applying transformer-based text summarization for
  keyphrase generation. Lobachevskii Journal of Mathematics  \textbf{44}(1),
  123--136 (2023). \doi{10.1134/S1995080223010134}

\bibitem{golovizninaautomatic}
Goloviznina, V., Kotelnikov, E.: Automatic summarization of {R}ussian texts:
  Comparison of extractive and abstractive methods. In: Computational
  Linguistics and Intellectual Technologies: Proceedings of the International
  Conference “Dialogue 2022”. pp. 223--235 (2022).
  \doi{10.28995/2075-7182-2022-21-223-235}

\bibitem{gupta2019abstractive}
Gupta, S., Gupta, S.K.: Abstractive summarization: An overview of the state of
  the art. Expert Systems with Applications  \textbf{121},  49--65 (2019).
  \doi{10.1016/j.eswa.2018.12.011}

\bibitem{hulth2003improved}
Hulth, A.: Improved automatic keyword extraction given more linguistic
  knowledge. In: Proceedings of the 2003 conference on Empirical methods in
  natural language processing. pp. 216--223 (2003).
  \doi{10.3115/1119355.1119383}

\bibitem{jiang2023generating}
Jiang, Y., Meng, R., Huang, Y., Lu, W., Liu, J.: Generating keyphrases for
  readers: A controllable keyphrase generation framework. Journal of the
  Association for Information Science and Technology  (2023).
  \doi{10.1002/asi.24749}

\bibitem{kilgarriff2001comparing}
Kilgarriff, A.: Comparing corpora. International Journal of Corpus Linguistics
  \textbf{6} (11 2001). \doi{10.1075/ijcl.6.1.05kil}

\bibitem{krapivin2009large}
Krapivin, M., Autaeu, A., Marchese, M.: Large dataset for keyphrases extraction
   (2009)

\bibitem{kulkarni2022learning}
Kulkarni, M., Mahata, D., Arora, R., Bhowmik, R.: Learning rich representation
  of keyphrases from text. In: Findings of the Association for Computational
  Linguistics: NAACL 2022. pp. 891--906 (2022).
  \doi{10.18653/v1/2022.findings-naacl.67}

\bibitem{lewis2020bart}
Lewis, M., Liu, Y., Goyal, N., Ghazvininejad, M., Mohamed, A., Levy, O.,
  Stoyanov, V., Zettlemoyer, L.: {BART}: Denoising sequence-to-sequence
  pre-training for natural language generation, translation, and comprehension.
  In: Proceedings of the 58th Annual Meeting of the Association for
  Computational Linguistics. pp. 7871--7880 (2020).
  \doi{10.18653/v1/2020.acl-main.703}

\bibitem{lin2004rouge}
Lin, C.Y.: {ROUGE}: A package for automatic evaluation of summaries. In: Text
  summarization branches out. pp. 74--81 (2004)

\bibitem{liu2019roberta}
Liu, Y., Ott, M., Goyal, N., Du, J., Joshi, M., Chen, D., Levy, O., Lewis, M.,
  Zettlemoyer, L., Stoyanov, V.: Ro{BERT}a: A robustly optimized {BERT}
  pretraining approach. arXiv preprint arXiv:1907.11692  (2019).
  \doi{10.48550/arXiv.1907.11692}

\bibitem{loshchilov2018decoupled}
Loshchilov, I., Hutter, F.: Decoupled weight decay regularization. In:
  International Conference on Learning Representations (2018)

\bibitem{malykh2021generating}
Malykh, V., Porplenko, D., Tutubalina, E.: Generating sport summaries: A case
  study for {R}ussian. In: Analysis of Images, Social Networks and Texts: 9th
  International Conference, AIST 2020, Skolkovo, Moscow, Russia, October
  15--16, 2020, Revised Selected Papers 9. pp. 149--161. Springer (2021).
  \doi{10.1007/978-3-030-72610-2_11}

\bibitem{meng2017deep}
Meng, R., Zhao, S., Han, S., He, D., Brusilovsky, P., Chi, Y.: Deep keyphrase
  generation. In: ACL 2017-55th Annual Meeting of the Association for
  Computational Linguistics, Proceedings of the Conference (Long Papers). pp.
  582--592 (2017). \doi{10.18653/v1/P17-1054}

\bibitem{miftahutdinov2020biomedical}
Miftahutdinov, Z., Alimova, I., Tutubalina, E.: On biomedical named entity
  recognition: experiments in interlingual transfer for clinical and social
  media texts. In: Advances in Information Retrieval: 42nd European Conference
  on IR Research, ECIR 2020, Lisbon, Portugal, April 14--17, 2020, Proceedings,
  Part II 42. pp. 281--288. Springer (2020). \doi{10.1007/978-3-030-45442-5_35}

\bibitem{rietzler2020adapt}
Rietzler, A., Stabinger, S., Opitz, P., Engl, S.: Adapt or get left behind:
  Domain adaptation through {BERT} language model finetuning for aspect-target
  sentiment classification. In: Proceedings of the 12th Language Resources and
  Evaluation Conference. pp. 4933--4941 (2020)

\bibitem{rubio2022hulat}
Rubio, A., Mart{\'\i}nez, P.: {HULAT-UC3M at SimpleText@ CLEF-2022: Scientific
  text simplification using BART}. Proceedings of the Working Notes of CLEF
  (2022)

\bibitem{schutz2008keyphrase}
Schutz, A.T.: Keyphrase extraction from single documents in the open domain
  exploiting linguistic and statistical methods  (2008)

\bibitem{shen2023enhanced}
Shen, L., Le, X.: An enhanced method on transformer-based model for one2seq
  keyphrase generation. Electronics  \textbf{12}(13), ~2968 (2023).
  \doi{10.3390/electronics12132968}

\bibitem{song2023survey}
Song, M., Feng, Y., Jing, L.: A survey on recent advances in keyphrase
  extraction from pre-trained language models. Findings of the Association for
  Computational Linguistics: EACL 2023 pp. 2108--2119 (2023).
  \doi{10.18653/v1/2023.findings-eacl.161}

\bibitem{swaminathan2020preliminary}
Swaminathan, A., Zhang, H., Mahata, D., Gosangi, R., Shah, R., Stent, A.: A
  preliminary exploration of {GAN}s for keyphrase generation. In: Proceedings
  of the 2020 Conference on Empirical Methods in Natural Language Processing
  (EMNLP). pp. 8021--8030 (2020). \doi{10.18653/v1/2020.emnlp-main.645}

\bibitem{syed2021menuner}
Syed, M.H., Chung, S.T.: Menu{NER}: Domain-adapted {BERT} based {NER} approach
  for a domain with limited dataset and its application to food menu domain.
  Applied Sciences  \textbf{11}(13), ~6007 (2021). \doi{10.3390/app11136007}

\bibitem{tank2022text}
Tank, M., Thakkar, P.: Text summarization approaches under transfer learning
  and domain adaptation settings—a survey. In: Computational Intelligence and
  Data Analytics: Proceedings of ICCIDA 2022, pp. 73--88. Springer (2022).
  \doi{10.1007/978-981-19-3391-2_5}

\bibitem{vaca2022extractive}
Vaca, A., Segurado, A., Betancur, D., Jim{\'e}nez, {\'A}.B.: Extractive and
  abstractive summarization methods for financial narrative summarization in
  {E}nglish, {S}panish and {G}reek. In: Proceedings of the 4th Financial
  Narrative Processing Workshop@ LREC2022. pp. 59--64 (2022)

\bibitem{wan2008single}
Wan, X., Xiao, J.: Single document keyphrase extraction using neighborhood
  knowledge. In: AAAI. vol.~8, pp. 855--860 (2008)

\bibitem{wang2022automatic}
Wang, S., Jiang, J., Huang, Y., Wang, Y.: Automatic keyphrase generation by
  incorporating dual copy mechanisms in sequence-to-sequence learning. In:
  Proceedings of the 29th International Conference on Computational
  Linguistics. pp. 2328--2338 (2022)

\bibitem{wright2022generating}
Wright, D., Wadden, D., Lo, K., Kuehl, B., Cohan, A., Augenstein, I., Wang,
  L.L.: Generating scientific claims for zero-shot scientific fact checking.
  In: Proceedings of the 60th Annual Meeting of the Association for
  Computational Linguistics (Volume 1: Long Papers). pp. 2448--2460 (2022).
  \doi{10.18653/v1/2022.acl-long.175}

\bibitem{wu2022pre}
Wu, D., Ahmad, W.U., Chang, K.W.: Pre-trained language models for keyphrase
  generation: A thorough empirical study. arXiv preprint arXiv:2212.10233
  (2022). \doi{10.48550/arXiv.2212.10233}

\bibitem{yadav2021forumbert}
Yadav, A., Milde, B.: forum{BERT}: Topic adaptation and classification of
  contextualized forum comments in {G}erman. In: Proceedings of the 17th
  Conference on Natural Language Processing (KONVENS 2021). pp. 193--202 (2021)

\bibitem{ye2018semi}
Ye, H., Wang, L.: Semi-supervised learning for neural keyphrase generation. In:
  Proceedings of the 2018 Conference on Empirical Methods in Natural Language
  Processing. pp. 4142--4153 (2018). \doi{10.18653/v1/D18-1447}

\bibitem{zhang2020pegasus}
Zhang, J., Zhao, Y., Saleh, M., Liu, P.J.: {PEGASUS}: pre-training with
  extracted gap-sentences for abstractive summarization. In: Proceedings of the
  37th International Conference on Machine Learning. pp. 11328--11339 (2020)

\bibitem{zhangbertscore}
Zhang, T., Kishore, V., Wu, F., Weinberger, K.Q., Artzi, Y.: {BERTS}core:
  Evaluating text generation with {BERT}. In: International Conference on
  Learning Representations

\bibitem{zmandar2023comparative}
Zmandar, N., El-Haj, M., Rayson, P.: A comparative study of evaluation metrics
  for long-document financial narrative summarization with transformers. In:
  International Conference on Applications of Natural Language to Information
  Systems. pp. 391--403. Springer (2023). \doi{10.1007/978-3-031-35320-8_28}

\bibitem{zolotareva2020abstractive}
Zolotareva, E., Tashu, T.M., Horv{\'a}th, T.: Abstractive text summarization
  using transfer learning. In: ITAT. pp. 75--80 (2020)

\end{thebibliography}

\end{document}